\documentclass{article}

\PassOptionsToPackage{numbers}{natbib}

\usepackage[preprint]{neurips_2024}

\usepackage[utf8]{inputenc} 
\usepackage[T1]{fontenc}    
\usepackage{hyperref}       
\usepackage{url}            
\usepackage{booktabs}       
\usepackage{amsfonts}       
\usepackage{nicefrac}       
\usepackage{microtype}      
\usepackage{xcolor}         
\usepackage{pifont}

\usepackage{graphicx}
\usepackage{booktabs} 
\usepackage{float}
\usepackage{floatflt}
\usepackage{amsmath}
\usepackage{amssymb}
\usepackage{mathtools}
\usepackage{amsthm}
\usepackage{bbm}
\usepackage{bm}
\usepackage{tikz}
\usepackage{ctable}
\usepackage{multicol}
\usepackage{multirow}
\usepackage{caption}
\usepackage{subcaption}
\usepackage{wrapfig}
\usepackage{algorithm}
\usepackage{algpseudocode}

\DeclareMathOperator*{\argmin}{arg\,min}
\DeclareMathOperator*{\argmax}{arg\,max}
\def\our{HyperMask}
\newcommand{\xmark}{\ding{55}}%

\title{HyperMask: Adaptive Hypernetwork-based Masks for Continual Learning}

\author{Kamil Książek$^{1,2}$ \quad Przemysław Spurek$^{2, 3}$ \\
$^1$ Institute of Theoretical and Applied Informatics, Polish Academy of Sciences, Gliwice, Poland \\
$^2$ Jagiellonian University, Faculty of Mathematics and Computer Science, Cracow, Poland \\ 
$^3$ IDEAS NCBR, Warsaw, Poland \\
\texttt{kamil.ksiazek@uj.edu.pl, przemyslaw.spurek@uj.edu.pl} \\
}

\begin{document}

\maketitle

\begin{abstract}
    Artificial neural networks suffer from catastrophic forgetting when they are sequentially trained on multiple tasks. Many continual learning (CL) strategies are trying to overcome this problem. One of the most effective is the hypernetwork-based approach. The hypernetwork generates the weights of a target model based on the task's identity. The model's main limitation is that, in practice, the hypernetwork can produce completely different architectures for subsequent tasks. 
    To solve such a problem, we use the lottery ticket hypothesis, which postulates the existence of sparse subnetworks, named winning tickets, that preserve the performance of a whole network. In the paper, we propose a method called \our{}, which dynamically filters a target network depending on the CL task. The hypernetwork produces semi-binary masks to obtain dedicated target subnetworks. 
    Moreover, due to the lottery ticket hypothesis, we can use a single network with weighted subnets. Depending on the task, the importance of some weights may be dynamically enhanced while others may be weakened. \our{} achieves competitive results in several CL datasets and, in some scenarios, goes beyond the state-of-the-art scores, both with derived and unknown task identities.
\end{abstract}

\section{Introduction}

Learning from a continuous data stream is challenging for deep learning models. Artificial neural networks suffer from catastrophic forgetting~\citep{mccloskey1989catastrophic} and drastically forget previously known information upon learning new knowledge.
Continual learning (CL)~\cite{hsu2018re} methods aim to learn consecutive tasks and prevent forgetting already learned ones effectively. 


One of the most promising approaches to CL is the hypernetwork~\cite{henning2021posterior,von2019continual} approach.
A hypernetwork architecture~\cite{ha2016hypernetworks} is a neural network that generates weights for a separate target network designated to solve a specific task.
In a continual learning setting, a hypernetwork generates the weights of a target model based on the task identity. 
At the end of training, we have a single meta-model, which produces dedicated weights. Due to the ability to generate completely different weights for each task, hypernetwork-based models feature minimal forgetting.  
Unfortunately, such properties were obtained by producing completely different architectures for substantial tasks. 
Only hypernetworks use information about tasks. Such a model can produce different nets for each task and solve them separately. The hypernetwork cannot use the weight of the target network from the previous task.

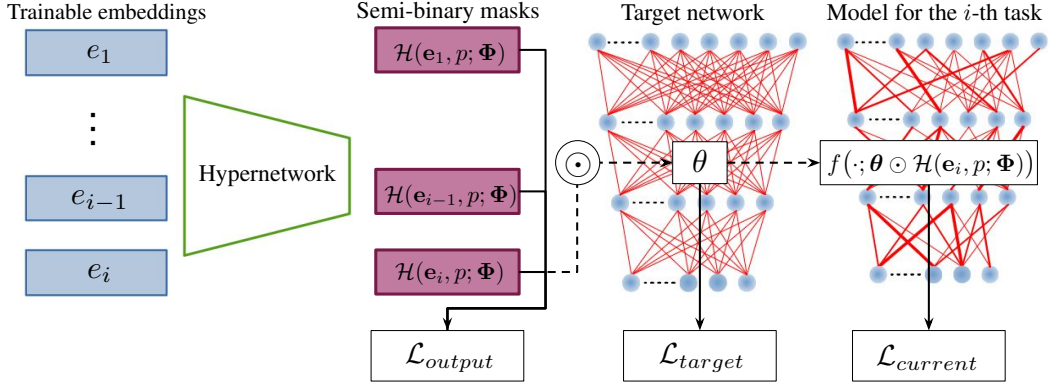
\begin{figure}[!h]
    \begin{center}

    \begin{tikzpicture}[scale=0.27]
    \node[inner sep=0pt] (russell) at (-5.0,0)
    {\includegraphics[trim={0,5cm, 0,5cm, 0,1cm, 0,8cm},clip,width=0.98\linewidth]{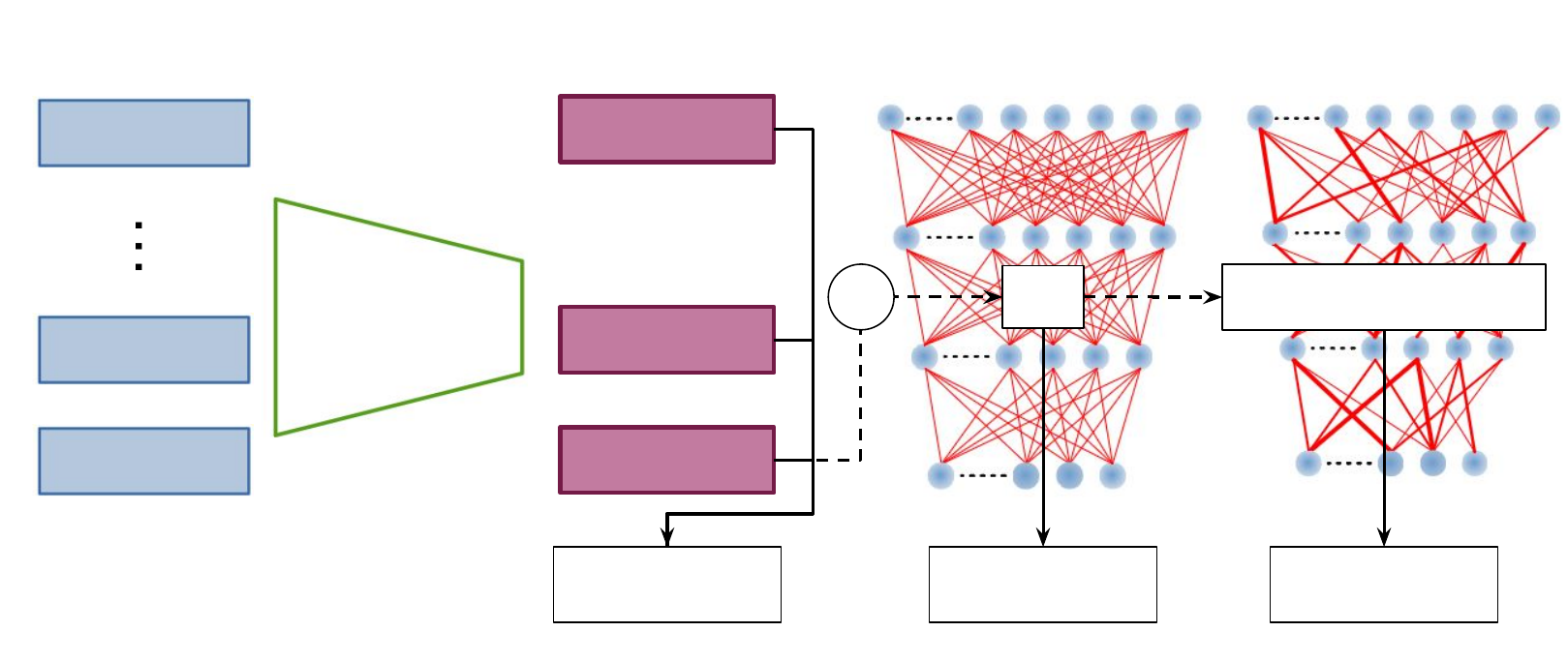} };
    \node[scale=1.1] at (-26.35,6.75) {$e_1$};
    \node[scale=1.2] at (-26.35,-0.5) {$e_{i-1}$};  
    \node[scale=1.2] at (-26.35,-4.05) {$e_{i}$};  
    
    \node[scale=0.9] at (-18.1,0.75) {Hypernetwork };
    \node[scale=0.85] at (-26.0,8.75) {Trainable embeddings };    
    \node[scale=0.9] at (-9.2,8.8) {Semi-binary masks };
    \node[scale=0.9] at (2.85,8.7) {Target network };
    \node[scale=0.9] at (14.75,8.8) {Model for the $i$-th task};
    \node[scale=1.4] at (-2.8,1.3) {$\odot$ };
    \node[scale=1.2] at (3.2,1.5) {\bf $\theta$ };

    \node[scale=0.88] at (-9.1,6.75) {$\mathcal{H}( \mathbf{e}_1, p ; \bm{\Phi} )$};
    \node[scale=0.88] at (-9.05,-0.25) {$\mathcal{H}( \mathbf{e}_{i-1}, p ; \bm{\Phi} )$};
    \node[scale=0.88] at (-9.1,-4.05) {$\mathcal{H}( \mathbf{e}_{i}, p ; \bm{\Phi} )$};
    \node[scale=0.9] at (14.6,1.3) {\bf $f\big( \cdot ;  \bm{\theta} \odot \mathcal{H}(\mathbf{e}_{i},p;\bm{\Phi})\big)$ };

    \node[scale=1.1] at (-9.1,-8.2) {$\mathcal{L}_{output}$};
    \node[scale=1.1] at (3.3,-8.2) {$\mathcal{L}_{target}$};
    \node[scale=1.1] at (14.4,-8.2) {$\mathcal{L}_{current}$};
    
    \end{tikzpicture}

    \end{center}
    \caption{Commonly, the parameters of a neural network are directly adjusted from data to solve a task. In \our{}, hypernetwork maps embedding vectors $e_i$ to the semi-binary mask, producing a subnetwork dedicated to the target network to solve the $i$--th task. Finally, a subset of weights was switched off (removed connections in the right-side image) while the remaining weights were scored continuously from 0 to 1 (weights of different thicknesses).}
    \label{fig:merf_appro}
\end{figure}

To solve such a problem, we use the lottery ticket hypothesis (LTH) \cite{frankle2018lottery}, which postulates that we can find subnetworks named winning tickets with performance similar (or even better) to the full architecture. However, the search for optimal winning tickets in continual learning scenarios is difficult \cite{mallya2018piggyback,wortsman2020supermasks}, as consecutive learning steps require repetitive pruning 
and retraining for each arriving task, which is impractical.
Alternatively, Winning SubNetworks (WSN) \cite{kang2022forget} incrementally learns model weights and task-adaptive binary masks. 
WSN eliminates catastrophic forgetting by freezing the subnetwork weights considered essential for the previous tasks and memorizing masks for all tasks.

Our paper proposes a method called \our{}\footnote{The source code is available at \url{https://github.com/gmum/HyperMask}} which combines hypernetwork and lottery ticket hypothesis paradigms.
Hypernetwork produces semi-binary masks to the target network to obtain weighted subnetworks dedicated to new tasks; see Fig.~\ref{fig:merf_appro}. By semi-binary masks, we mean that some weights may be turned off for selected tasks while the remaining weights are scored between 0 and 1. The masks produced by the hypernetwork modulate the weights of the main network and act like dynamic filters, enhancing the important target weights for a given task and decreasing the importance of the remaining weights. Additionally, a subset of weights may be excluded from contribution in a given task by sparsifying the mask. Consequently, we work on a single network with subnetworks dedicated to each task. When \our{} learns a new task, we reuse the learned subnetwork weights from the previous tasks. Also, we can simultaneously train both the hyper- and the target network, or just train the hypernetwork with frozen target network weights after initialization.
\our{} also inherits the ability of the hypernetwork to adapt to new tasks with minimal forgetting. We produce a semi-binary mask directly from the trained task embedding vector, which creates a dedicated subnetwork for each dataset.

To the best of our knowledge, our model is the first architecture-based CL model that uses hypernetwork, or, in general, any meta-model, for producing masks for other networks. Updates of hypernetworks are prepared not directly for the weights of the main model, like in~\cite{von2019continual}, but for masks dynamically filtering the target model.

Our contributions can be summarized as follows:
\begin{itemize}
    \item We propose a method using the hypernetwork paradigms for modelling the lottery ticket-based subnetwork. The hypernetwork modulates the weights of the main model instead of their direct preparation as in~\cite{von2019continual}. 
    \item We show that \our{} can reuse weights from the lottery ticket module and adapt to new tasks from the hypernetwork paradigm. 
    \item We demonstrate that \our{} achieves competitive results for the presented CL datasets and obtains state-of-the-art scores for some demanding scenarios, both with known and unknown task identities.
\end{itemize}

\section{Related Works}

\paragraph{Continual learning} Typically, continual learning approaches are divided into three main categories: regularization, dynamic architectures, and replay-based techniques \citep{de2021continual,parisi2019continual,wang2023comprehensive}. \our{} model belongs to the dynamic architectures category.

Architecture-based approaches use dynamic architectures that dedicate separate model branches to different tasks. These branches can be developed incrementally, such as in the case of Progressive Neural Networks \cite{rusu2016progressive}. 
The architecture of such a system can be optimized to enhance parameter effectiveness and knowledge transfer, for example, by reinforcement learning (RCL \citep{xu2018reinforced}, BNS \citep{qin2021bns}), architecture search (LtG \citep{li2019learn}, BNS \citep{qin2021bns}), and variational Bayesian methods (BSA \citep{kumar2021bayesian}).
Alternatively, a static architecture can be reused with iterative pruning as proposed by PackNet \citep{mallya2018packnet} or by applying Supermasks~\citep{wortsman2020supermasks}.

In PackNet~\cite{mallya2018packnet}, during the training of consecutive tasks, a subset of available network weights is selected while the rest is pruned. Furthermore, weights considered important for some tasks must remain unchanged and cannot be pruned in the future. Therefore, the space for modifying weights is getting smaller along with subsequent tasks. In~\our{}, we can create any number of embeddings and different semi-binary masks so that the target network weights can be filtered dynamically. These weights are also continuously scored and do not have to be switched on or off, like in PackNet.

Recent works are focused on efficiently using of a feature extractor fixed after the first learning stage. FeTrIL \cite{petit2023fetril} proposes a new approach which combines a fixed feature extractor and a pseudo-features generator. The generator uses geometric translations to produce a favourable representation of old classes based on new classes. In FeCAM \cite{goswami2023fecam}, a fixed feature extractor is used to build separate representations of elements in each class. Then, the class prototype and a corresponding covariance matrix are estimated. The final predictions are obtained by the calculation of the Mahalanobis distance between samples and different class representations. Both methods work in class incremental settings.

\paragraph{Hypernetworks for continual learning}

A hypernetwork~\cite{ha2016hypernetworks} is an architecture which generates a vector of weights for a separate target network designated to solve a specific task. Hypernetworks are widely used, e.g. in generative models \cite{spurek2020hypernetwork}, building implicit  representation~\cite{szatkowski2023hypernetworks} and few-shot learning \cite{sendera2023hypershot}.

In a continuous learning environment, a hypernetwork generates the weights of a target model based on the task's identity. 
HNET~\cite{von2019continual} uses task embeddings to produce weights dedicated to each task.
HNET can be seen as an architecture-based strategy as we create a distinct architecture for each task. However, it can also be viewed as a regularization model. After training, a single meta-model producing specialized weights is left. 
Hypernetworks can generate different nets for tasks and, in some sense, solve them independently. While in HNET, a meta-model generates target network weights directly, in \our{}, a meta-model prepares masks for the target network adapted for consecutive tasks. In~\cite{henning2021posterior}, authors propose a Bayesian version of the hypernetworks in which they produce parameters of the prior distribution of the Bayesian network. 

\paragraph{Pruning-based Continual Learning}

Most architecture-based methods use additional memory to obtain
better performance. In the pruning-based method, we build computationally- and memory-efficient strategies.

CLNP \citep{golkar2019continual} freezes the most significant parts of the network and divides it into active, inactive and interference parts. This pruning approach is based on neuron average activity. Therefore, the model maintained its performance after learning of the following tasks. However, the current model's performance and capacity for the upgoing tasks must be compromised.

Piggyback \citep{mallya2018piggyback} uses a pre-trained model and task-specific binary masks. In contrast, \our{} for all the target network weights considered important, create their continuous score between 0 and 1. Also, in PiggyBack, the base network weights are frozen while \our{} freezing the target network is optional and its effectiveness depends on the considered model.

HAT \citep{serra2018overcoming} uses task-specific learnable attention vectors to recognize significant weights for each task and dynamically creates or destroys paths between single network units. Furthermore, a cumulative attention vector is created. In \our{}, masks are produced by a meta-model and depend on the trained embedding vectors. Data-free Subnetworks (DSN)~\cite{gao2024enhancing} is a modification of HAT that adds components designed for forward and backward transfer. The mechanism uses data-free replay. The model is different from ours since the authors do not use meta-models. The authors use the term \emph{hypernetwork} to refer to the main model used for pruning. In our paper, the hypernetwork produces weight to another network, called the target network, according to~\cite{ha2016hypernetworks}.

LL-Tickets \citep{chen2020long} show that we can find a subnetwork, referred to as lifelong tickets, which performs well on all tasks during continual learning. If the tickets cannot work on the new task, the method looks for more prominent tickets from the existing ones.
However, the LL-Tickets expansion process is made up of a
series of retraining and pruning steps. 

In Winning SubNetworks (WSN) \cite{kang2022forget}, authors propose to jointly learn the model and task-adaptive binary masks dedicated to task-specific subnetworks (winning tickets).
Unfortunately, WSN eliminates catastrophic forgetting by freezing the subnetwork weights for the previous tasks and memorizing masks for all tasks. Meanwhile, \our{} dynamically enhances important target weights for a given task and decreases the significance of the remaining weights. A modification of WSN, i.e. Soft-SubNetworks (SoftNet)~\cite{kang2023softnet}, uses non-binary masks, but this is a method designed for few-shot incremental learning (each class is represented only by a few training samples), while HyperMask is adapted for many-shot incremental learning in which all tasks and classes are equally important.



\section{\our{}: Adaptive Hypernetworks for Continual Learning}\label{sec:HyperMask}

Our proposition, called \our{}, is a hypernetwork-based continual learning method. In \our{}, the hypernetwork returns semi-binary masks to produce weighted target subnetworks dedicated to new tasks. 
This solution inherits the ability of the hypernetwork to adapt to new tasks with minimal forgetting and uses the lottery ticket hypothesis to create a single network with weighted masks.

\paragraph{Problem statement}

Let us consider a supervised learning setup where $T$ tasks are derived to a learner sequentially. We assume that $X_t = \{  \bm{x}_{i,t} \}^{n_t}_{i=1} $ is the dataset for task $t$,
composed of $n_t$ elements of raw instances and $Y_t = \{ y_{i,t} \}^{n_t}_{i=1} $ are the corresponding labels, $t \in \lbrace 1, 2, ..., T \rbrace$. Data from task $t$ we denote by $D_t = (X_{t}, Y_{t})~\subset~X~\times~Y$.
We assume a neural network $f( \cdot ;   \bm{\theta})$, parameterized
by the model weights $\bm{\theta}$ and the standard continual learning scenario
\begin{equation*}
  \bm{\theta}^{*} = \argmin\limits_{\bm{\theta}} \frac{1}{n_t} \sum_{i=1}^{n_t} \mathcal{L} \big( f(\bm{x_{i,t}} ;   \bm{\theta}) \big),
\end{equation*}
where $\mathcal{L}(\cdot, \cdot)$ is a classification objective loss such as the cross-entropy loss. We assume that $D_t$ for task $t$ is only accessible when learning task $t$. 
Also, the task identity is given in both the training and testing stages, except for the additional series of experiments described further.

To provide space for learning future tasks, a continuing learner often adopts over-parameterized deep neural networks. In such situations, we can find subnets with equal or better performance.

\paragraph{Hypernetwork}

Hypernetworks, introduced in \cite{ha2016hypernetworks}, are defined as neural models that generate weights for a separate target network solving a specific task. 
Before we present our solution, we describe the classical approach to using hypernetworks in CL.
In this case, a hypernetwork generates individual weights for subsequent CL tasks. 
In HNET~\cite{henning2021posterior,von2019continual} the authors propose using trainable embeddings $  \bm{e}_t \in \mathbb{R}^{N}$, where $t \in \{ 1, ..., T \}$. The hypernetwork $\mathcal{H}$ with weights $\bm{\Phi}$ generates weights $ \bm{\theta}_t$ of the target network $f$ dedicated to the $t$--th task
$$  
\mathcal{H}( \bm{e}_t ; \bm{\Phi} ) = \bm{\theta}_t.
$$
HNET meta-architecture (hypernetwork) produces different weights for each continual learning task.
We have the function $  f_{\bm{\theta}_t} : X \to Y$ (a neural network classifier with weights $  \bm{\theta}_t$), which predicts labels based on data samples from a continuous learning dataset. 

The target network is not trained directly.   
In HNET, the authors use a hypernetwork \linebreak
$H_{\bm{\Phi}}: \mathbb{R}^N \ni \bm{e}_t  \to \bm{\theta}_t$,
which for a task embedding $  \bm{e}_t$ returns weights $ \bm{\theta}_t$ of the corresponding target network $  f_{\bm{\theta}_t}: X \to Y$.
Thus, each continual learning task is represented by a classifier function 
\begin{equation*}
f_{\bm{\theta}_t} = f(\cdot;\bm{\theta}_t) = f\big( \cdot ; H(\bm{e}_t ; \bm{\Phi})\big).
\end{equation*}
At the end of training, we have a single meta-model, which produces dedicated weights. Due to the ability to generate completely different weights for each task, hypernetwork-based models feature minimal forgetting.  

\subsection{\our{} -- overview }
In \our{}, a hypernetwork $\mathcal{H}_{\bm{\Phi}}: \mathbb{R}^N \ni \mathbf{e}_t  \to \bm{m}_t,$ where $ t \in \{ 1, ..., T \}$, produces semi-binary masks which are multiplied element-wise with the target network weights. Similarly, as in HNET, we use trainable embeddings $\bm{e}_t$ to prepare masks adapted to a given task. 
\paragraph{Mask creation} To ensure that a mask has a continuous representation ranging from 0 to 1, we use the {\em tanh} activation function on the hypernetwork output. Then, we select $(1 - p)\%$ weights with the highest weight scores, where $(1 - p)$ is the ratio of the target layer capacity and $c(p, i, t; \mathbf{x})$ is a threshold value for the selected target layer $\mathbf{x}$, during the $i$-th training iteration of the $t$-th task. The choice of weights is performed through the task-dependent semi-binary weight mask $\bm{m}_t$, where an absolute value greater than the threshold denotes that the weight is considered during the forward pass and is zeroed otherwise. Formally, $\bm{m}_t$ is obtained by applying an indicator function $\sigma_p(\cdot;\cdot)$ to a weight $w$ being an element of $\mathbf{x}$ which represents a single target network layer being the output of $\mathcal{H}$:
\begin{equation*} 
   \sigma_p( w; \mathbf{x} )  =\left\{
  \begin{array}{@{}ll@{}}
    0 & \text{if}\  |w| \leq c(p, i, t; \mathbf{x}),\\
    w & \text{otherwise}.
  \end{array}\right.
\end{equation*}
The mask sparsifying is done before the element-wise multiplication with the target network weights. Additionally, the ratio $p$ is constant starting from the second task but, for the first trained task, is gradually increased from $0$ to $p$
\begin{equation*}   
    c(p, i, t; \mathbf{x})  =\left\{
  \begin{array}{@{}ll@{}}
    P(p; |\mathbf{x}|) & \text{if}\  t > 1,\\
    P(\frac{i}{n} \cdot p; |\mathbf{x}|) & \text{otherwise}.
  \end{array}\right.
\end{equation*}
$P(p; |\mathbf{x}|)$ represents the $p$-th percentile of the set of absolute values of a given mask layer. Each task is trained through $n$ iterations. The absolute value of consecutive $\mathbf{x}$ weights is calculated element-wise. 

Therefore, \our{} uses trainable embeddings $\mathbf{e}_t \in \mathbb{R}^{N}$ for $t \in \{ 1, ..., T \}$, threshold level $p$ and hypernetwork $\mathcal{H}$ with weights $  \bm{\Phi}$ generating a semi-binary mask $\bm{m}_t$ (with $p\%$ zeros) adapted to each task and applied for the target network weights $\bm{\theta}$
\begin{equation*}     
\mathcal{H}( \mathbf{e}_t, p ; \bm{\Phi} ) = \sigma_{p}\big(:, \mathcal{H}( \mathbf{e}_t ; \bm{\Phi} )\big) = \bm{m}_t;
\end{equation*}
$\sigma_{p}(:,\cdot)$ means that the indicator function is applied for all output values of $\mathcal{H}$.

\paragraph{Target network} In \our{}, we can train both the hyper- and target network, or just the hypernetwork and this process is governed by the hyperparameter $\bm{\theta}_{train}$. When the target weights $\bm{\theta}$ are fixed, the hypernetwork modulates $\bm{\theta}$ values being constant after initialization.

In the opposite case, we have two trainable architectures. The hypernetwork $\bm{\Phi}$ and the target network weights $\bm{\theta}$ are simultaneously trained with a cross-entropy loss function. 
More precisely, we model the classifier function $f_{\bm{\theta}}: X \to Y$ with general weights $\bm{\theta}$ used in all tasks. Thus, each continual learning task is represented by
\begin{equation*}
f(\cdot;  \bm{\theta} \odot \bm{m}_t ) = f\big( \cdot ;  \bm{\theta} \odot \mathcal{H}(\mathbf{e}_t,p;\bm{\Phi})\big),
\end{equation*}
where $\odot$ is element-wise multiplication. 


In the training procedure, we have added two regularization terms. The first one is the output regularizer proposed by~\cite{li2017learning}:
$$  
\mathcal{L}_{regularizer} = \sum_{t=1}^{T-1} \sum_{i=1}^{| X_{t} |} \| f(\mathbf{x}_{i,t} ;  \bm{\theta}^{*} \odot \bm{m}^{*}_t) - f(\mathbf{x}_{i,t} ;  \bm{\theta} \odot \bm{m}_t) \|^2,
$$
where $\bm{\theta}^{*}$ and $\bm{m}^{*}_t$ are the target network parameters and mask designed for task $t$ before attempting to learn task $T$, respectively. This solution is expensive in terms of memory and does not follow the online learning paradigm adequately. However, hypernetworks \cite{henning2021posterior,von2019continual} avoid this problem. Task-conditioned hypernetworks produce an output depending on the task embedding.  
We can compare the fixed hypernetwork output produced before learning task $T$ with weights $\bm{\Phi}^{*}$, with the output after a current proposition of hypernetwork weight modifications $  \Delta \bm{\Phi}$, according to the cross-entropy loss. Finally, in \our{}, the output regularization loss is given by:
$$
\mathcal{L}_{output} (\bm{\Phi}^{*}, \bm{\Phi}, \Delta \bm{\Phi}, \{ \mathbf{e}_t\}  ) =  \frac{1}{T-1} \sum_{t=1}^{T-1} \| \mathcal{H}(\mathbf{e}_{t}, 0; \bm{\Phi}^{*}) - \mathcal{H}(\mathbf{e}_{t}, 0; \bm{\Phi} + \Delta \bm{\Phi}) \|^2, 
$$
where $\Delta \bm{\Phi}$ is an update proposition to the hypernetwork weights. We do not sparse the hypernetwork weights at this stage, therefore $p=0$.

The difference between \our{} and~\cite{von2019continual} relies on the fact that using $\mathcal{L}_{output}$, we just regularize masks dedicated to consecutive continual learning tasks and not the target weights. When target weights are trainable, we include classical $L_1$ regularization designed for them
$$  
\mathcal{L}_{target}( \bm{\theta}^{*}_t,  \bm{\theta}_t) = \|  \bm{\theta}^{*}_t -  \bm{\theta}_t \|_{1}, 
$$
where $   \bm{\theta}^{*}_t$ is the set of target network parameters before attempting to learn task $T$. Optionally, we can multiply $ 
\mathcal{L}_{target}$ by the hypernetwork-generated mask (masked $L_1$) to ensure that the most important target network weights will not be drastically modified while the other ones will be more susceptible to adjustments. In such a case
\begin{align*}
\mathcal{L}_{target}( \bm{\theta}^{*}_t,  \bm{\theta}_t, \bm{m}_t) = \bm{m}_t \odot \|  \bm{\theta}^{*}_t -  \bm{\theta}_t \|_{1}.
\end{align*}
During hyperparameter optimization, we compared two variants of $ 
\mathcal{L}_{target}$, i.e. masked and non-masked $L_1$. A conclusive choice is dependent on the considered dataset.

\begin{table*}[!h]
    \begin{minipage}{0.65\textwidth}
    \centering
    \caption{Results of different continual learning methods, in terms of mean overall accuracy with a standard deviation. HyperMask-F and HyperMask-T denote fixed or trainable weights of the target network, respectively. The target architecture for Permuted and Split MNIST datasets was MLP, while for Split CIFAR-100 and Tiny ImageNet, it was ResNet-18.
    Results for different methods than \our{} are derived from other papers, except for WSN, for which we additionally calculated experiments with Split CIFAR-100 and Tiny ImageNet using ResNet-18, to ensure a fair comparison with \our{}. Results for the other WSN architectures are presented in Appendix~\ref{app:known_task_ID}. Details about experimental scenarios, and \our{} hyperparameters are presented in Appendix~\ref{app:architectures}. We obtained the best results for CIFAR-100 while for Permuted MNIST, HyperMask was better than its primary baselines: WSN and HNET. }
    \centering
    {
     \scriptsize
     \begin{tabular}{@{}l@{}c@{}c@{\;}c@{\;}c@{}}
        \FL
    	\textbf{Method} & \textbf{Permuted MNIST} & \textbf{Split MNIST} & \textbf{Split CIFAR-100} & \textbf{Tiny ImageNet}\ML
        HAT & $\mathbf{97.67 \pm 0.02}$ & $-$ &$72.06 \pm 0.50$ & $-$ \\
        GPM & $94.96 \pm 0.07$ & $-$ &$73.18 \pm 0.52$ & $67.39 \pm 0.47$\\
        PackNet & $96.37 \pm 0.04$ & $-$ &$72.39 \pm 0.37$ & $55.46 \pm 1.22$\\
        SupSup & $96.31 \pm 0.09$ & $-$ & $75.47 \pm 0.30$ & $59.60 \pm 1.05$\\
        La-MaML & $-$ & $-$ & $71.37 \pm 0.67$ & $66.99 \pm 1.65$\\ 
        FS-DGPM & $-$ & $-$ & $74.33 \pm 0.31$ & $70.41 \pm 1.30$\ML
        WSN, $c = 50\%$ & $96.24 \pm 0.11$ & $-$ & $73.29 \pm 0.43$ & $\mathbf{79.84 \pm 0.56}$\ML
        EWC & $95.96 \pm 0.06$ & $99.12 \pm 0.11$ & $72.77 \pm 0.45$ & $-$\\
        SI & $94.75 \pm 0.14$ & $99.09 \pm 0.15$ & $-$ & $-$\\
        DGR & $97.51 \pm 0.01$ & $99.61 \pm 0.02$ & $-$ & $-$\\
        HNET+ENT & $97.57 \pm 0.02$ & $\mathbf{99.79 \pm 0.01}$ & $-$ & $-$ \ML

        \our{}-F (our) & $90.77 \pm 0.06$ & $99.35 \pm 0.04$ & $\mathbf{81.36 \pm 0.57}$ & $74.90 \pm 0.81$\\

        \our{}-T (our) & $97.66 \pm 0.04$ & $99.64 \pm 0.07$ & $76.49 \pm 1.28$ & $74.23 \pm 0.85$ \LL
        
    \end{tabular}
    }
    \label{tab:continual}
    \end{minipage}
    \begin{minipage}{0.34\textwidth}
        \includegraphics[trim={0,2cm, 0,2cm, 0,2cm, 0,2cm},clip,width=\linewidth]{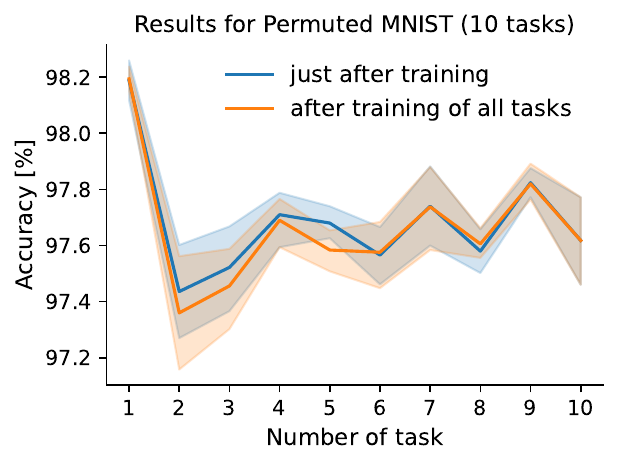}
        \includegraphics[trim={0,2cm, 0,2cm, 0,2cm, 0,2cm},clip,width=\linewidth]{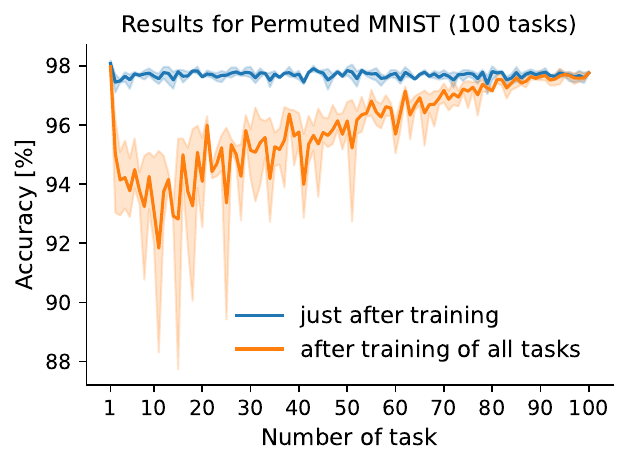}
        \captionof{figure}{Visualization of mean accuracy (with 95\% confidence intervals) for Permuted MNIST for 10 and 100 tasks. The blue lines represent test accuracy calculated after training consecutive models, while the orange lines correspond to test accuracy after finishing all CL tasks. The mean accuracy of the Permuted MNIST 100-task case equals $95.92 \pm 0.18$.\label{permuted:best_set_10_tasks}}
    
    \end{minipage}
\end{table*}

The final cost function consists of the classical cross-entropy $\mathcal{L}_{current}$, output regularization $\mathcal{L}_{output}$, and, when the target network parameters are trainable, also target regularization $\mathcal{L}_{target}$ 
$$
\mathcal{L} = \mathcal{L}_{current} + \beta \cdot \mathcal{L}_{output} + \mathbbm{1}(\bm{\theta}_{train} \mathrm{~is~True}) \cdot \lambda \cdot \mathcal{L}_{target}, 
$$
where $\beta$ and $\lambda$ are hyperparameters which control the strength of regularization and $\mathbbm{1}(\cdot)$ is an indicator function. It is necessary to emphasize that both regularization components are crucial when we have two trainable networks because we have to prevent radical changes in the weights of both networks after learning of subsequent CL tasks. The first regularization component $\mathcal{L}_{output}$ is responsible for the regularization of the hypernetwork weights producing masks, while the second one, $\mathcal{L}_{target}$, accounts for the target network weights. In~\cite{von2019continual}, hypernetworks directly produce the target model’s weights. Thus, additional regularization is redundant.

The pipeline of \our{} is summarized in Algorithm~\ref{alg:hypermask} presented in Appendix~\ref{app:pseudocode}.



\section{Experiments}
\label{experiments}
This section presents a numerical comparison of our model with a few baseline solutions. We analyzed task-incremental continual learning with a multi-head setup for all the experiments. We followed the experimental setups from recent works \cite{deng2021flattening,goswami2023fecam,saha2020gradient,yoon2020scalable}.

\paragraph{Architectures}
As target networks, we used two-layered MLP with 1000 neurons per layer for Permuted MNIST and Split MNIST. For Split CIFAR-100 and Tiny ImageNet, we selected the ResNet-18 architecture~\cite{he2015ResNet}. In all cases, hypernetworks were MLPs with one or two hidden layers. A detailed description of the architectures, selected hyperparameters and their optimization are presented in Appendix~\ref{app:architectures}. 

\paragraph{Baselines} We compared our solution with two natural baselines: WSN \cite{kang2022forget} and HNET \cite{von2019continual}. WSN used the lottery ticket hypothesis, while HNET used the hypernetwork paradigm.
We also added a comparison with strong CL baselines from different categories. In particular, we selected regularisation-based methods: HAT \cite{serra2018overcoming} and EWC \cite{kirkpatrick2017overcoming}, rehearsal-based methods like GPM~\cite{saha2020gradient} and FS-DGPM~\cite{deng2021flattening}, a pruning-based method like PackNet \cite{mallya2018packnet} and SupSup \cite{wortsman2020supermasks}, and a meta learning approach like La-MAML~\cite{gupta2020lamaml}. In CL settings without known task identities, we also compared with the state-of-the-art exemplar-free class incremental learning methods like FeTrIL~\cite{petit2023fetril} and FeCAM~\cite{goswami2023fecam}. 

\paragraph{Experimental settings} 
For the Permuted MNIST and Split MNIST datasets, we have chosen the protocol from HNET~\cite{von2019continual}, while for Tiny ImageNet~\cite{le2015tiny}, we selected the setting from WSN. In the case of Split CIFAR-100, with known task identities, we divided the dataset into 10 tasks with ten classes, as in WSN~\cite{kang2022forget}. For Split CIFAR-100 with unknown task identities, we divided the dataset into 5 tasks with 20 classes, like in FeCAM~\cite{goswami2023fecam}. Some results in the tables were directly taken from papers. To ensure a fair comparison with \our{}, we additionally performed experiments for WSN using the ResNet-18 architecture for Split CIFAR-100 and Tiny ImageNet.

\paragraph{Results for known task identity scenarios}
We evaluated our algorithm on four standard benchmark datasets: Permuted MNIST, Split MNIST, Split CIFAR-100, and Tiny ImageNet. In Table~\ref{tab:continual}, we present a summary of the results with known task identity while its full version is inserted in Table~\ref{app:known_task_ID}in Appendix~\ref{app:known_task_ID}. Results for \our{} are depicted in two variants: \our{}-F with a fixed target network model, and \our{}-T with trainable two networks. In a demanding Split CIFAR-100 scenario, we obtained better results than all reference methods for both \our{} approaches. Moreover, we had the second score in Permuted MNIST and Split MNIST for a trainable variant. In the case of Permuted MNIST, our exact result was equal to 97.664, so it was only 0.006 smaller than HAT. We also observed that \our{} with a trainable target model was very efficient when the target network was MLP. When it was a convolutional architecture, \our{} with frozen target weights was more promising. However, for Tiny ImageNet the differences between results for two \our{} variants were much smaller than for Split CIFAR-100.
\begin{figure}[!h]
    \centering
            \includegraphics[trim={0,4cm, 0,4cm, 0,6cm, 0,3cm},clip,width=0.495\linewidth]{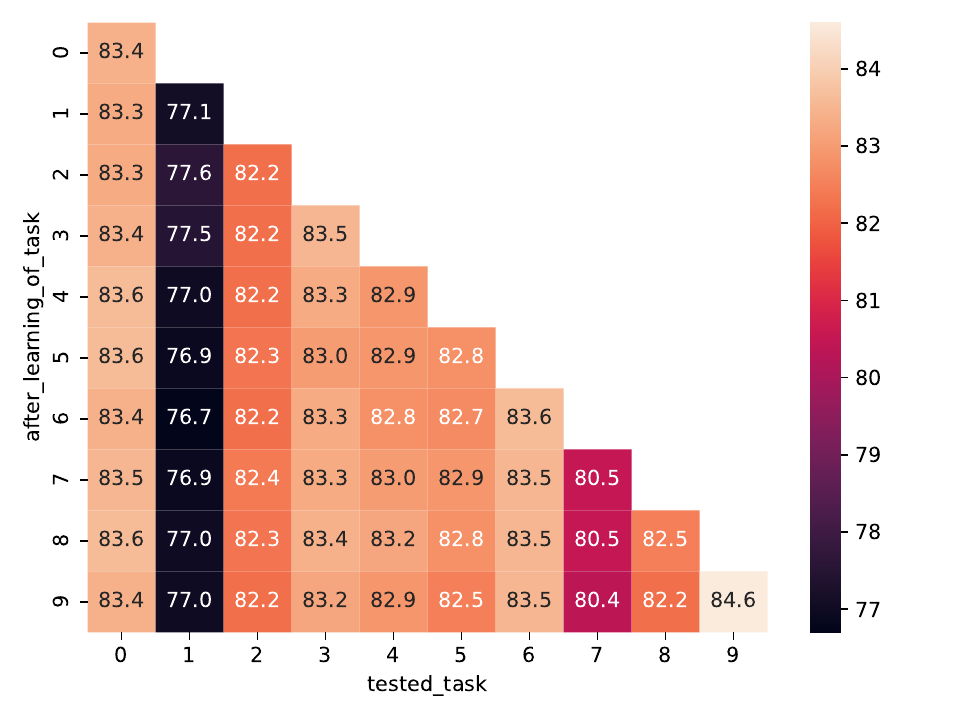}
            \includegraphics[trim={0,4cm, 0,4cm, 0,6cm, 0,3cm},clip,width=0.495\linewidth]{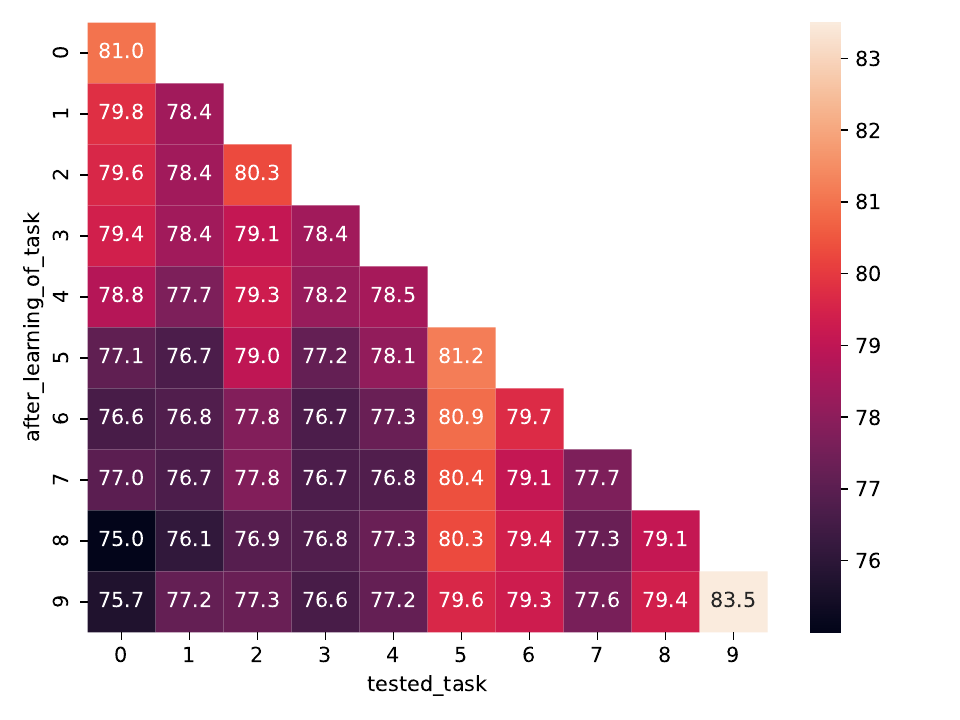}
    \caption{Mean test accuracy for consecutive CL tasks averaged over five runs of \our{} for the fixed (left side) or trainable (right side) ResNet-18 target network architectures for ten tasks of the CIFAR-100 dataset. For the fixed target model, only a slight decrease in overall accuracy for consecutive tasks was noticed, while for the trainable target model, the decline was more significant (for instance, from $81.0$ to $75.7 \%$ for the first CL task). Also, mean results after all CL tasks were higher for the fixed target network scenario.\label{cifar_100:accuracy_matrix}}
\end{figure}
\begin{wrapfigure}{L}{0.5\textwidth}
    \centering
            \includegraphics[trim={0,2cm, 1,1cm, 0,2cm, 0,25cm},clip,width=\linewidth]{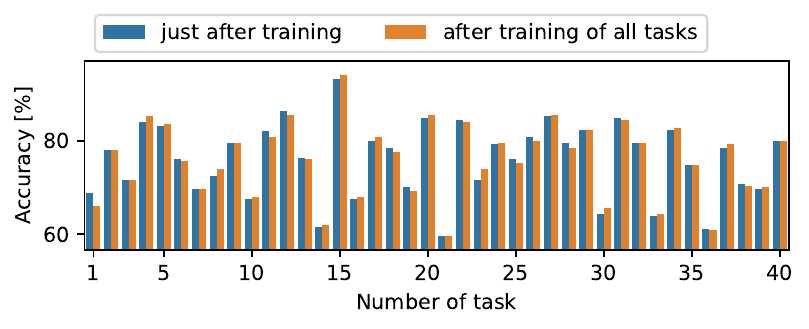}
            \includegraphics[trim={0,2cm, 0,3cm, 0,2cm, 1,025cm},clip,width=\linewidth]{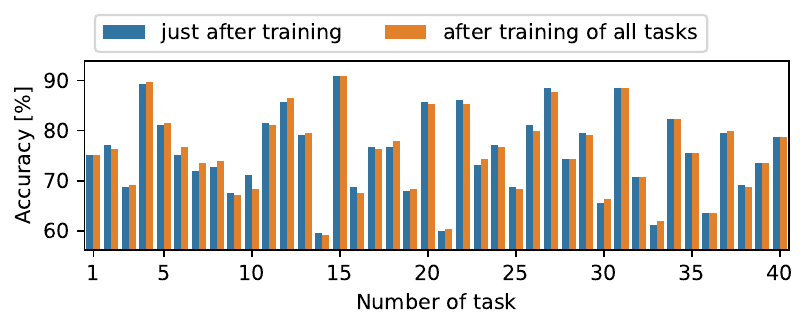}
    \caption{The accuracy of \our{} for two models with the ResNet-18 architecture, trained through 40 tasks on Tiny ImageNet. The first row presents results for \our{}-F, with an accuracy of $76.04\%$, while the second row depicts scores of \our{}-T, with an accuracy of $75.54\%$. Appendix~\ref{app:backward} shows the results for all ten models.\label{tinyimagenet:accuracy_shortcut}}
\end{wrapfigure}
\paragraph{Forgetting of previous tasks}
In Fig.~\ref{permuted:best_set_10_tasks} we present mean accuracy (with 95\% confidence intervals) of \our{}-T in 10 and 100 task scenarios of Permuted MNIST dataset, with MLPs as target models. Interestingly, \our{} maintains accuracy on the first task, even after training on many subsequent ones.  It may indicate that the trainable hyper- and target MLP tandem adapts to the first task, affecting further weight behaviour.

In Fig.~\ref{cifar_100:accuracy_matrix} we compare the test accuracy of the two considered \our{} variants (with ResNet-18 target models) after training of consecutive CL tasks for Split CIFAR-100 with known task identity. \our{}-F obtained a very slight negative backward transfer but the decrease in accuracy of \our{}-T was more severe. 

We also examined \our{} in terms of previous tasks' forgetting for Tiny ImageNet. Fig.~\ref{tinyimagenet:accuracy_shortcut} depicts the test accuracies for subsequent CL tasks for two selected models. The negative backward transfer was not substantial, and for some CL tasks, the accuracy even increased after gaining knowledge from the following tasks. The mean backward transfer for Tiny ImageNet was $-0.06 \pm 0.09$ for HyperMask-F and $-0.11 \pm 0.07$ for HyperMask-T. A more detailed analysis of the specific cases of backward transfer, as well as aggregated results after training consecutive tasks, are presented in Appendix~\ref{app:backward}.

\paragraph{Stability of \our{} model}

\begin{wrapfigure}{L}{0.7\textwidth}
    \centering
            \includegraphics[trim={0,2cm, 0,3cm, 0,2cm, 0,35cm},clip,width=0.48\linewidth]{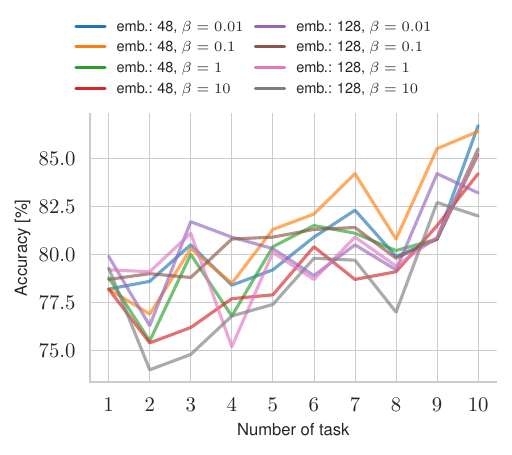}%
            \includegraphics[trim={0,2cm, 0,3cm, 0,2cm, 0,35cm},clip,width=0.52\linewidth]{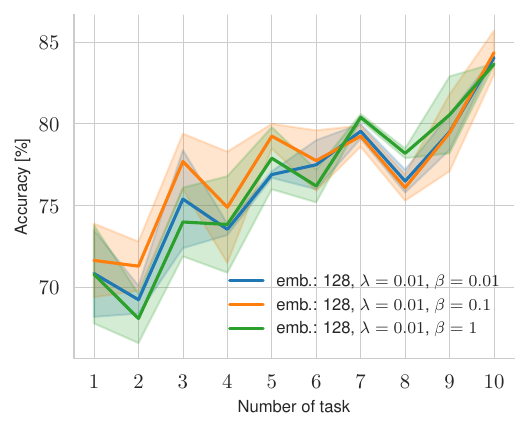}
    \caption{The results of consecutive CL tasks for Split CIFAR-100, for \our{}-F (left-hand side, single training run) and \our{}-T (right-hand side, two training runs). For both cases, $p$ was set to 0, and a hypernetwork contained two hidden layers with 100 neurons.
    \label{cifar:ablation}}
\end{wrapfigure}

\our{} models have a similar number of hyperparameters as HNET. The most critical parameters are $\beta$ and $\lambda$, which control regularization strength.
We also use a hyperparameter describing the level of zeros in a semi-binary mask, $p$, and define whether masked or non-masked $L_1$ must be used. Masked $L_1$ means that $\mathcal{L}_{target}$ was multiplied by the hypernetwork-generated mask while non-masked $L_1$ is the opposite case. We performed an extensive ablation study for different datasets and \our{} settings. Its results are presented in Appendix~\ref{app:ablation} while scores for various \our{} configurations for Split CIFAR-100 are depicted in Fig.~\ref{cifar:ablation}.


\paragraph{Results for unknown task identity scenarios}
We also evaluated \our{} in a scenario in which task identity is not directly given to the model but must be inferred by the network itself. Following~\cite{von2019continual}, we prepared a task inference method based on the entropy values (\our{} + ENT). After training for all tasks, consecutive data samples were propagated through the hyper- and target network for different task embeddings. The task with the lowest entropy value of the classification layer's output in the target network was selected for the final calculations. Then, the classifier decision for the corresponding embedding was considered. We also proposed a hybrid of \our{} and FeCAM~\cite{goswami2023fecam} in which \our{} prepares features extraction and FeCAM creates task prototypes and selects a class based on \our{} output. Initially, FeCAM worked on a feature extractor trained only with classes present in the first CL task, while in our case, FeCAM uses the output of \our{} trained sequentially in all CL tasks.

\begin{table*}[t]
    \centering
    \caption{Mean overall accuracy (in $\%$) in a scenario where the model must recognize task identity. 
    The Split CIFAR-100 results for two \our{} variants are presented for the best model (max) and after averaging over five runs to enable a fair comparison with other methods.
    Other results are derived from HNET~\cite{von2019continual}, marked with $^\star$, and FeCAM~\cite{goswami2023fecam}, marked with $^{\star \star}$.}
    \scriptsize
      \begin{tabular}{@{}p{2,6cm}p{1,6cm}p{1,55cm}|p{0,75cm}p{2cm}p{2cm}}
    	\FL
     \multirow{2}{*}{\textbf{Method}} & \multirow{2}{*}{\textbf{Permuted MNIST}} & \multirow{2}{*}{\textbf{Split MNIST}} & \multicolumn{3}{c}{\textbf{Split CIFAR-100}} \\ 
     & & & & {\tiny last task accuracy} & {\tiny average task accuracy} \ML
        HNET+ENT$^\star$ & $91.75 \pm 0.21$ & $69.48 \pm 0.80$ & & $-$ & $-$\\
        EWC$^\star$ & $33.88 \pm 0.49$ & $19.96 \pm 0.07$ & & $-$ & $-$ \\
        SI$^\star$ & $29.31 \pm 0.62$ & $19.99 \pm 0.06$ & & $-$ & $-$ \\
        DGR$^\star$ & $96.38 \pm 0.03$ & $\mathbf{91.79 \pm 0.32}$ & & $-$ & $-$ \ML
        Euclidean-NCM$^{\star \star}$ & $-$ & $-$ & & $30.6$ & $50.0$ \\
        FeTrIL$^{\star \star}$ & $-$ & $-$ & & $46.2$ & $61.3$ \\
        FeCAM$^{\star \star}$ & $-$ & $-$ & & $48.1$ & $62.3$ \ML

        \our{}+ENT (our) & \multirow{2}{*}{$-$} & \multirow{2}{*}{$-$} & (max) & $46.3$ & $59.5$ \\
        {\tiny (fixed target)} & & & (5 runs) & $45.8 \pm 0.4$ & $59.0 \pm 0.5$ \\
        \our{}+FeCAM (our) & \multirow{2}{*}{$-$} & \multirow{2}{*}{$-$} & (max) & $\mathbf{55.9}$ & $\mathbf{66.2}$ \\
        {\tiny (fixed target)} & & & (5 runs) & $\mathbf{55.6 \pm 0.6}$ & $\mathbf{65.5 \pm 0.7}$ \ML
        
        \our{}+ENT (our) & \multirow{2}{*}{$89.43 \pm 1.49$} & \multirow{2}{*}{$62.89 \pm 5.83$} & (max) & $42.2$ & $54.9$ \\
        {\tiny (trainable target)} & & & (5 runs) & $41.4 \pm 0.5$ & $53.3 \pm 1.1$ \\
        \our{}+FeCAM (our) & \multirow{2}{*}{$\mathbf{97.03 \pm 0.07}$} & \multirow{2}{*}{$83.28 \pm 3.25$} & (max) & $49.8$ & $61.7$ \\
        {\tiny (trainable target)} & & & (5 runs) & $49.2 
        \pm 0.5$ & $60.0 \pm 1.0$ \LL
     \end{tabular}
    \label{tab:CL3}
\end{table*}


Table~\ref{tab:CL3} presents results for three datasets: Permuted MNIST (10 tasks), Split MNIST and CIFAR-100 (5 tasks per 20 classes, like in Table 5 in~\cite{goswami2023fecam}). Even \our{} + ENT achieved very competitive results, but \our{} + FeCAM (with a trainable target model) outperformed all baselines in Permuted MNIST. For a more demanding scenario with CIFAR-100, \our{} + FeCAM (with a fixed target model) obtained very high accuracy compared to other methods. Both the last and average task accuracies were higher than in the case of FeTrIL and FeCAM. Furthermore, even a more straightforward method, i.e. \our{} + ENT, achieved comparable performance to other solutions. WSN~\cite{kang2022forget} only realizes the strategy in which the task identity is known in advance, and the authors did not describe in the paper a method for task inference. Therefore, we did not evaluate WSN in such a scenario.
A detailed description of the \our{} settings and considered scenarios are available in Appendix~\ref{app:architectures}. 

\paragraph{Limitations and future works}
One of the main limitations of \our{} is the memory consumption because the hypernetwork output layer must have the same number of neurons as the number of parameters in the target network. The chunking approach described in~\cite{von2019continual}, in which the target's weight values are generated by the hypernetwork partially, was not adopted in \our{} because it led to considerably worse results so far. However, this approach should be analyzed thoroughly. It should facilitate taking advantage of larger architectures like Vision Transformers. Also, one has to consider a separation of hyper- and target network training strategies, for instance, with different learning rate settings for two architectures.


\section{Conclusion}

We present \our{}, a method that trains a hypernetwork producing semi-binary masks to generate target subnetworks tailored to new tasks. This approach utilizes the hypernetwork's capacity to adjust to new tasks with minimal forgetting. Also, due to the lottery ticket hypothesis, we can use a single network with weighted subnets devoted to each task. The experimental section shows that our model obtains very competitive results compared to recent baselines and, in some cases, outperforms the state-of-the-art methods. \our{} also has a potential for application for strategies in which task identity has to be inferred by the method.  




\bibliography{icml2024_conference}
\bibliographystyle{abbrvnat}


\appendix

\section{Appendix: Pseudocode of \our{}}\label{app:pseudocode}
The Algorithm~\ref{alg:hypermask} described the main steps of \our{}, described in detail in Section~\ref{sec:HyperMask}. Depending on the selected mode, the algorithm returns both updated hyper- and target network weights, $\bm{\Phi}$ and $\bm{\Theta}$, respectively, or just updated hypernetwork weights $\bm{\Phi}$.
\begin{algorithm}[!t]
    \caption{The pseudocode of \our{}.\label{alg:hypermask}}
    \begin{algorithmic}
    	\Require hypernetwork $\mathcal{H}$ with weights $\bm{\Phi}$, target network $f$ with weights $  \bm{\theta}$, sparsity $p \geq 0$, regularization strength $\beta > 0$, and $\lambda > 0$, $n$ training iterations, datasets $\lbrace D_1, D_2, ..., D_T\rbrace$, $  (\bm{x}_{i,t}, y_{i,t}) \in D_{t}, t \in \lbrace 1, ..., T \rbrace$, target network training mode $\bm{\theta}_{train}$
    	\Ensure updated hypernetwork weights $  \bm{\Phi}$, potentially with updated target network weights $  \bm{\theta}$
    \State Initialize randomly weights $  \bm{\Phi}$ and $ \bm{\theta}$ with embeddings $  \lbrace \bm{e}_1, \bm{e}_2, ..., \bm{e}_T \rbrace$
    	\For{$t \leftarrow 1$ to $T$}
    		\If{$t > 1$}
                \If{$\bm{\theta}_{train}$ is True}
                    \State $\bm{\theta}^{*} \leftarrow  \bm{\theta}$
                \EndIf
                \For{$t^{\prime} \leftarrow 1$ to $t-1$}
                    \State Store $\bm{m}_{t^{\prime}} \leftarrow \mathcal{H}(\bm{e}_{t^{\prime}}, p; \bm{\Phi})$
                \EndFor
            \EndIf
            \For{$i \leftarrow 1$ to $n$}
                \State $\bm{m}_t \leftarrow \mathcal{H}(\bm{e}_t, p; \bm{\Phi})$
                \State $\bm{\theta}_t \leftarrow \bm{m}_t \odot  \bm{\theta}$
                \State $\hat{y}_{i, t} \leftarrow f(\bm{x}_{i, t};  \bm{\theta}_t)$
                \If{$t = 1$}
                    \State $\mathcal{L} \leftarrow \mathcal{L}_{current}$\;
                \Else
                    \If{$\bm{\theta}_{train}$ is True}
                        \State $\mathcal{L} \leftarrow \mathcal{L}_{current} + \beta \cdot \mathcal{L}_{output} + \lambda \cdot \mathcal{L}_{target}$\;
                    \Else
                        \State $\mathcal{L} \leftarrow \mathcal{L}_{current} + \beta \cdot \mathcal{L}_{output}$\;
                    \EndIf
                \EndIf
                \If{$\bm{\theta}_{train}$ is True}
                    \State Update $\bm{\Phi}$ and $\bm{\theta}$
                \Else
                    \State Update $\bm{\Phi}$
                \EndIf
             \EndFor
            \State Store $\bm{e}_t$
    	\EndFor
    \end{algorithmic}
    \end{algorithm}

\section{Appendix: Full results for known task identity scenarios}\label{app:known_task_ID}
In Table~\ref{app:tab:full_continual}, we depict the extended results for known identity task scenarios. Previously, in Table~\ref{tab:continual}, for Split CIFAR-100 and Tiny ImageNet, we only presented the WSN results for ResNet-18, to ensure a fair comparison with \our{}. Now, in the case of WSN~\cite{kang2022forget}, we complement Split CIFAR-100 results for AlexNet, and Tiny ImageNet scores for TinyNet, as they were presented in the original paper. Although AlexNet achieved higher performance for Split CIFAR-100 than ResNet-18, it was still lower than \our{}, even with the trainable target architecture. For Tiny ImageNet, the results shown by the authors were worse than for ResNet-18.  

\begin{table*}[!h]
    \begin{minipage}{1.0\textwidth}
        \centering
        \caption{Full results of different continual learning methods, in terms of mean overall accuracy with corresponding standard deviations. HyperMask-F and HyperMask-T denote fixed or trainable weights of the target network, respectively. The target architecture for Permuted and Split MNIST datasets was MLP, while for Split CIFAR-100 and Tiny ImageNet, it was ResNet-18. Results for different methods than \our{} are derived from other papers, except for WSN, for which we additionally calculated experiments for Split CIFAR-100 and Tiny ImageNet using ResNet-18, to ensure a fair comparison with \our{}. We also inserted the original WSN results for AlexNet and TinyNet architectures. Details about experimental scenarios, and \our{} hyperparameters are presented in Appendix~\ref{app:architectures}. We obtained the best results for CIFAR-100 while for Permuted MNIST, HyperMask was better than its primary baselines: WSN and HNET.}
        \centering
        {
         \scriptsize
         \begin{tabular}{@{}l@{}c@{}>{\centering}p{1,75cm}|c@{\;}>{\centering}p{1,75cm}|c@{\;}c@{}}
            \FL
        	\textbf{Method} & \textbf{Permuted MNIST} & \textbf{Split MNIST} & \multicolumn{2}{c}{\textbf{Split CIFAR-100}} & \multicolumn{2}{c}{\textbf{Tiny ImageNet}}\ML
            HAT & $\mathbf{97.67 \pm 0.02}$ & $-$ & & $72.06 \pm 0.50$ & & $-$ \\
            GPM & $94.96 \pm 0.07$ & $-$ & & $73.18 \pm 0.52$ & & $67.39 \pm 0.47$\\
            PackNet & $96.37 \pm 0.04$ & $-$ & & $72.39 \pm 0.37$ & & $55.46 \pm 1.22$\\
            SupSup & $96.31 \pm 0.09$ & $-$ & & $75.47 \pm 0.30$ & & $59.60 \pm 1.05$\\
            La-MaML & $-$ & $-$ & & $71.37 \pm 0.67$ & & $66.99 \pm 1.65$\\ 
            FS-DGPM & $-$ & $-$ & & $74.33 \pm 0.31$ & & $70.41 \pm 1.30$\ML
            WSN, $c = 3\%$ & $94.84 \pm 0.11$ & $-$ & (AlexNet) & $70.65 \pm 0.36$ & (TinyNet) & $68.72 \pm 1.63$\\
            WSN, $c = 5\%$ & $95.65 \pm 0.03$ & $-$ & (AlexNet) & $72.44 \pm 0.27$ & (TinyNet) & $71.22 \pm 0.94$\\
            WSN, $c = 10\%$ & $96.14 \pm 0.03$ & $-$ & (AlexNet) & $74.55 \pm 0.47$ & (TinyNet) & $71.96 \pm 1.41$\\
            WSN, $c = 30\%$ & $96.41 \pm 0.07$ & $-$ & (AlexNet) & $75.98 \pm 0.68$ & (TinyNet) & $70.92 \pm 1.37$\\
            \multirow{2}{*}{WSN, $c = 50\%$} & \multirow{2}{*}{$96.24 \pm 0.11$} & \multirow{2}{*}{$-$} & (AlexNet) & $76.38 \pm 0.34$ & (TinyNet) & $69.06 \pm 0.82$\\
            & & & (ResNet-18) & $73.29 \pm 0.43$ & (ResNet-18) & $\mathbf{79.84 \pm 0.56}$\\
            WSN, $c = 70\%$ & $96.29 \pm 0.00$ & $-$ & & $-$ & & $-$\ML
            EWC & $95.96 \pm 0.06$ & $99.12 \pm 0.11$ & & $72.77 \pm 0.45$ & & $-$\\
            SI & $94.75 \pm 0.14$ & $99.09 \pm 0.15$ & & $-$ & & $-$\\
            DGR & $97.51 \pm 0.01$ & $99.61 \pm 0.02$ & & $-$ & & $-$\\
            HNET+ENT & $97.57 \pm 0.02$ & $\mathbf{99.79 \pm 0.01}$ & & $-$ & & $-$ \ML
    
            \our{}-F (our) & $90.77 \pm 0.06$ & $99.35 \pm 0.04$ & (ResNet-18) & $\mathbf{81.36 \pm 0.57}$ & (ResNet-18) & $74.90 \pm 0.81$\\
    
            \our{}-T (our) & $97.66 \pm 0.04$ & $99.64 \pm 0.07$ & (ResNet-18) & $76.49 \pm 1.28$ & (ResNet-18) & $74.23 \pm 0.85$ \LL
            
        \end{tabular}
        }
        \label{app:tab:full_continual}
        \end{minipage}
\end{table*}

\section{Appendix: Backward transfer }\label{app:backward}
HNET models produce completely different weights for each task. In consequence, they demonstrate minimal forgetting. \our{} models inherit such ability to minimize forgetting previous tasks thanks to masks created by hypernetworks. To measure the rate at which models forget previous tasks, we calculated a backward transfer (BWT) for the selected CL methods: WSN~\cite{kang2022forget}, HNET~\cite{von2019continual} and \our{}. BWT measures forgetting previous tasks after learning the last one:
\begin{equation*}
BWT = \frac{1}{T-1}\sum^{T-1}_{i=1}  A_{T ,i} - A_{i,i},
\end{equation*}
where $A_{T ,i}$ is the test overall accuracy for task $i$ after training on task $T$, while $A_{i, i}$ is the test overall accuracy for task $i$ just after training the model on this task. Negative BWT means that learning new tasks caused the forgetting of past tasks. Zero BWT represents a situation where the accuracy of CL tasks did not change after gaining new knowledge.
Finally, positive BWT corresponds to the state in which the model improved the accuracy of the previous CL tasks after learning the next ones.

Table~\ref{tab:transfer} presents mean backward transfer for five training runs of HNET+ENT and \our{}, for five experimental scenarios: Permuted MNIST with 10 or 100 tasks (for 100 tasks, three training runs were performed), Split MNIST, Split CIFAR-100 and Tiny ImageNet datasets. By definition, WSN remembers masks from the preceding tasks. Therefore, the backward transfer, in this case, is always equal to zero. HNET and \our{} achieved comparable and slightly negative values of BWTs for Permuted MNIST with 10 tasks and Split MNIST. In the case of Permuted MNIST with 100 CL tasks, we only have results for \our{}. Despite a much larger number of tasks, the negative backward transfer did not exceed $2\%$. 

Interestingly, for Split CIFAR-100, \our{} with fixed target network models achieved positive backward transfer which means that knowledge from the following tasks yielded slightly enhanced classification accuracy for the preceding tasks. In the case of trainable target models, the negative backward transfer exceeded $2.4\%$. For a demanding Tiny ImageNet, both \our{} variants suffered only slightly from the negative backward transfer. The presented results suggest that \our{} is largely immune to catastrophic forgetting.

\ctable[
caption = {Mean backward transfer (in $\%$) with standard deviation for different continual learning methods. \our{}-F denotes fixed while \our{}-T trainable target network models.},
label = tab:transfer,
mincapwidth = \textwidth,
doinside = {\scriptsize},
star = False,
pos = h
]{lccccc}{}{\FL
	\textbf{Dataset} & \multicolumn{2}{c}{\textbf{Permuted MNIST}} & \textbf{Split MNIST} & \textbf{Split CIFAR-100} & \textbf{Tiny ImageNet}\\
    & \textbf{10 tasks} & \textbf{100 tasks} & & & \ML
    WSN, $c = 30\%$ & $0.0$ & $-$ & $-$ & $0.0$ & $0.0$\\
    HNET+ENT & $-0.018 \pm 0.01$ & $-$ & $-0.027 \pm 0.07$ & $-$ & $-$ \ML
    \our{}-F (our) & $-$ & $-$ & $-$ & $0.031 \pm 0.19$ & $-0.059 \pm 0.09$ \\
    \our{}-T (our) & $-0.025 \pm 0.03$ & $-1.791 \pm 0.18$ & $-0.009 \pm 0.04$ & $-2.420 \pm 0.65$ & $-0.115 \pm 0.07$ \LL
}

\begin{figure*}[!t]
    \centering
        \begin{subfigure}{0.49\linewidth}
            \includegraphics[trim={0,2cm, 0,2cm, 0,2cm, 0,2cm},clip,width=\linewidth]{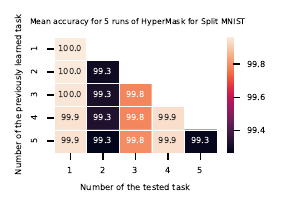}
        \end{subfigure}%
        \begin{subfigure}{0.49\linewidth}
            \includegraphics[trim={0,2cm, 0,2cm, 0,2cm, 0,2cm},clip,width=\linewidth]{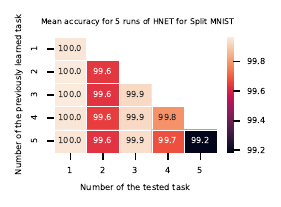}
        \end{subfigure}
    \caption{Mean test accuracy for consecutive CL tasks averaged over five runs of the best architecture settings of \our{} (left side) and the default setting of HNET (right side) for 5 tasks of the Split MNIST dataset. The performance drops after consecutive CL tasks are only slight. Interestingly, the second and the last tasks (i.e., classification of numbers 2 and 3 or 8 and 9, respectively) are the most challenging for both methods. \label{split:accuracy_matrix}}
\end{figure*}

\begin{figure}[!ht]
    \centering
        \begin{subfigure}{0.45\linewidth}
            \includegraphics[trim={0,2cm, 0,2cm, 0,2cm, 0,2cm},clip,width=\linewidth]{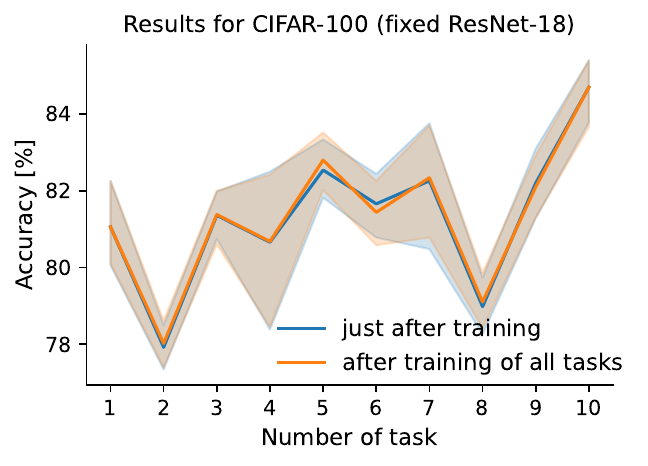}
        \end{subfigure}%
        \begin{subfigure}{0.45\linewidth}
            \includegraphics[trim={0,2cm, 0,2cm, 0,2cm, 0,2cm},clip,width=\linewidth]{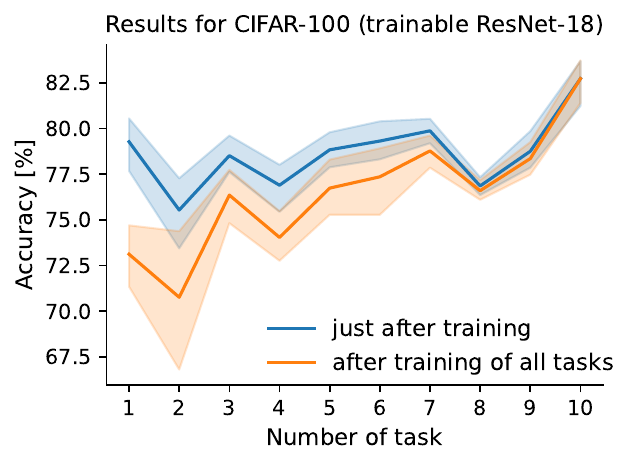}
        \end{subfigure}
    \caption{Visualization of mean accuracy (with 95$\%$ confidence intervals) of \our{} for 10 tasks of the CIFAR-100 dataset for the fixed (left side) and trainable (right side) ResNet-18 target network architecture. The blue lines represent test accuracy calculated after training subsequent models, while the orange lines correspond to test accuracy after finishing training for all CL tasks. One can conclude that the model with fixed ResNet-18 not only achieves higher classification accuracy but also less suffer from catastrophic forgetting than the trainable ResNet-18.\label{cifar:accuracy)tasks}}
\end{figure}


\begin{figure}
    \centering
            \includegraphics[trim={0,2cm, 0,2cm, 0,2cm, 0,2cm},clip,width=0.495\linewidth]{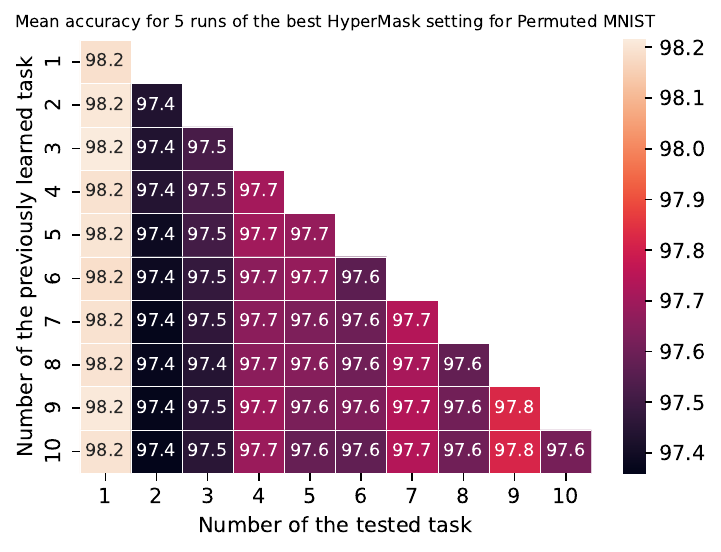}
            \includegraphics[trim={0,2cm, 0,2cm, 0,2cm, 0,2cm},clip,width=0.495\linewidth]{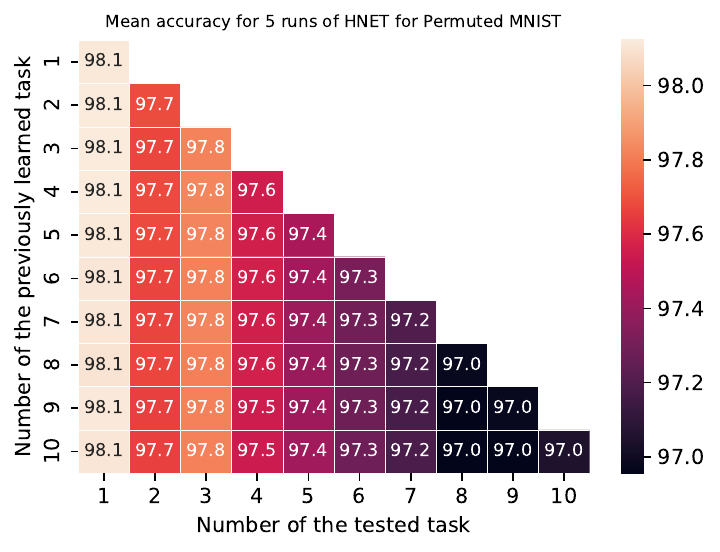}
    \caption{Mean test accuracy for consecutive CL tasks averaged over five runs of the best architecture settings of \our{} (left side) and the default setting of HNET (right side) for ten tasks of Permuted MNIST dataset. Training of subsequent tasks leads to a slight decrease in the overall accuracy of the previous tasks, but, in general, \our{} achieves higher accuracy for more recent tasks. However, HNET is more powerful for the first tasks.\label{permuted_10:accuracy_matrix}}
\end{figure}

\begin{figure}
    \begin{minipage}[t]{0.47\textwidth}
            \includegraphics[trim={0,25cm, 0,3cm, 0,28cm, 0,25cm},clip,width=0.95\linewidth]{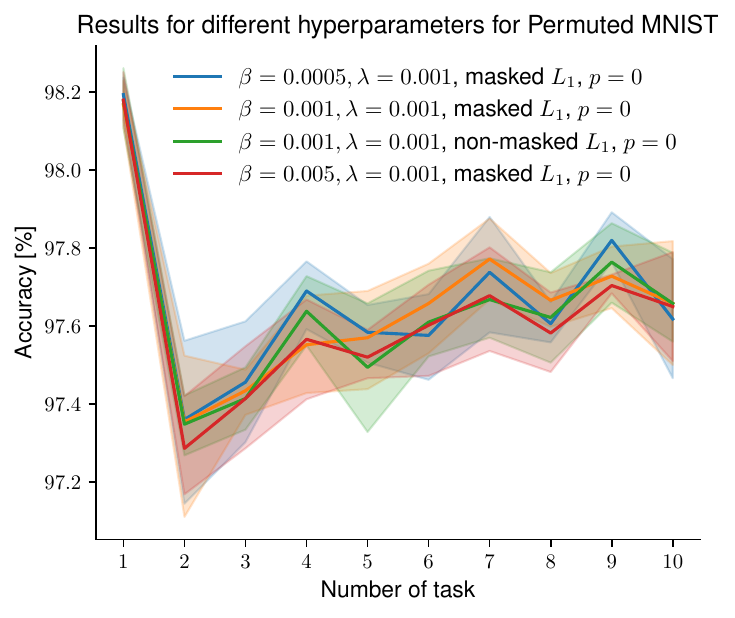}
            \caption{Visualization of stability of \our{} for Permuted MNIST. We obtained similar results for different hyperparameters.\label{permuted:more_settings}}
    \end{minipage}
    \begin{minipage}[t]{0.52\textwidth}
            \includegraphics[trim={0,2cm, 0,2cm, 0,2cm, 0,2cm},clip,width=0.9\linewidth]{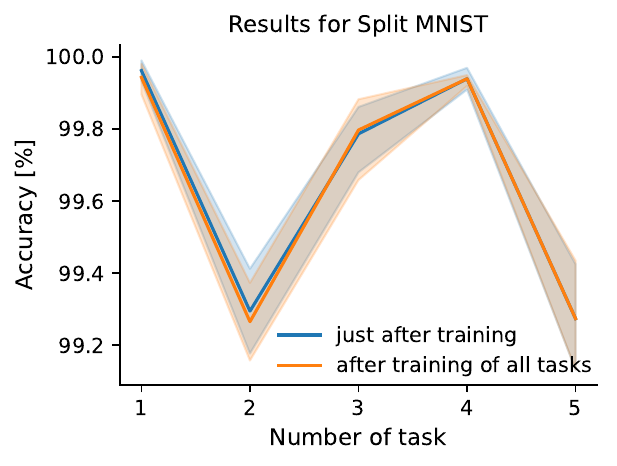}
            \caption{Visualization of mean accuracy (with 95\% confidence intervals) for Split MNIST for 5 tasks. The blue lines represent test accuracy calculated after training consecutive models, while the orange lines correspond to test accuracy after finishing all CL tasks. The decrease in accuracy after consecutive CL tasks is minimal.\label{split:5_tasks_training}}
    \end{minipage}
\end{figure}

\begin{figure*}[!h]
    \centering
       \begin{subfigure}{\linewidth}
            \includegraphics[trim={2,25cm, 10,8cm, 2,cm, 0,2cm},clip,width=0.5\linewidth]{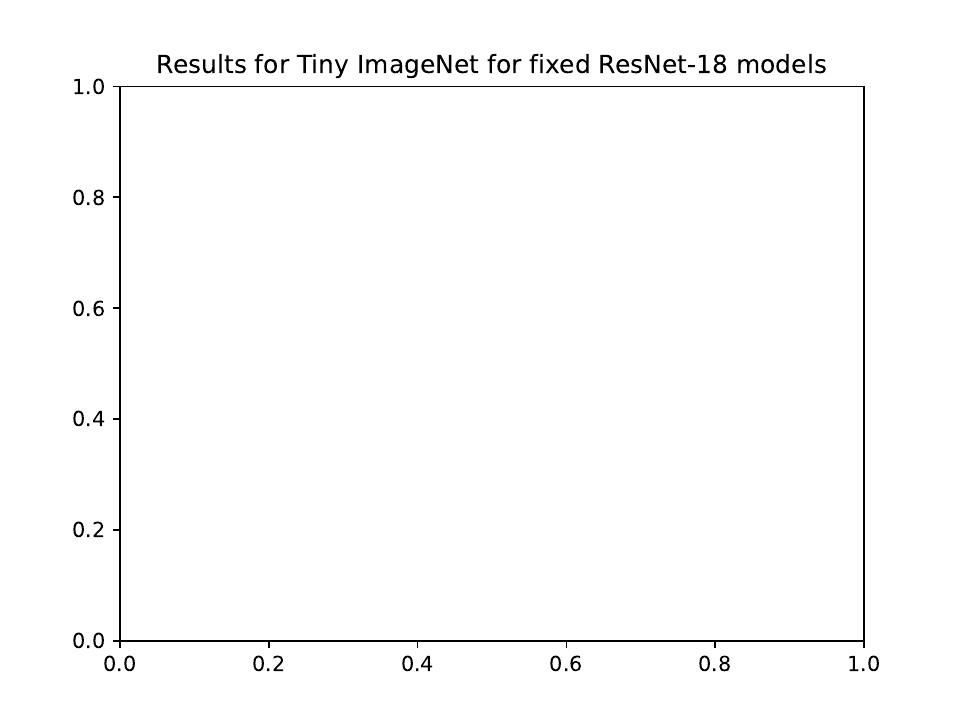}%
            \includegraphics[trim={2cm, 10,8cm, 1,6cm, 0,2cm},clip,width=0.5\linewidth]{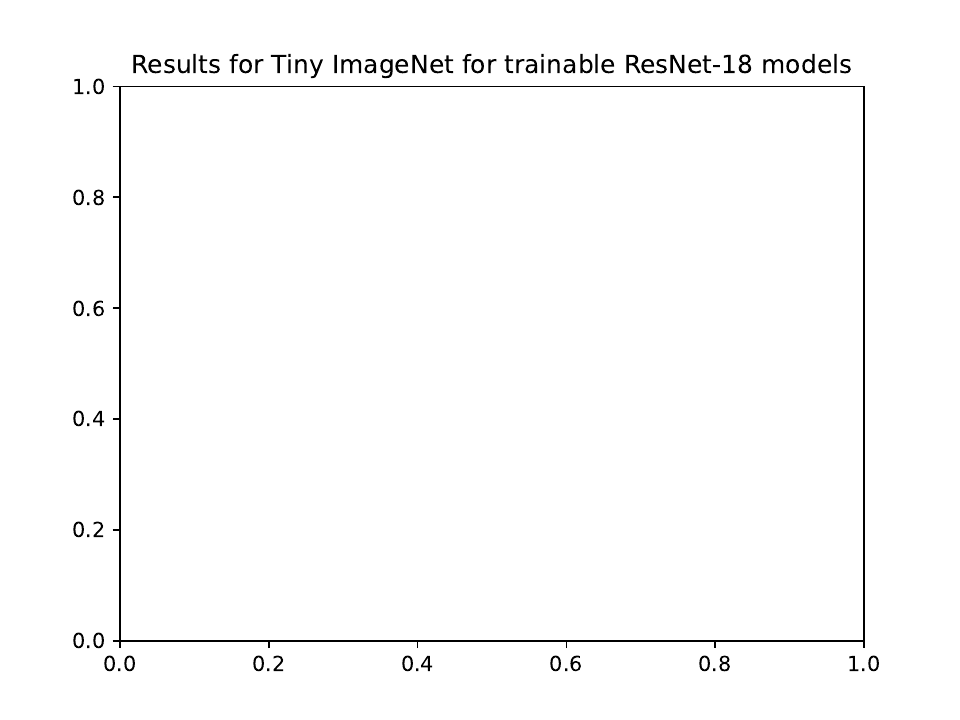}
            \includegraphics[trim={0,2cm, 1,1cm, 0,2cm, 0,2cm},clip,width=0.5\linewidth]{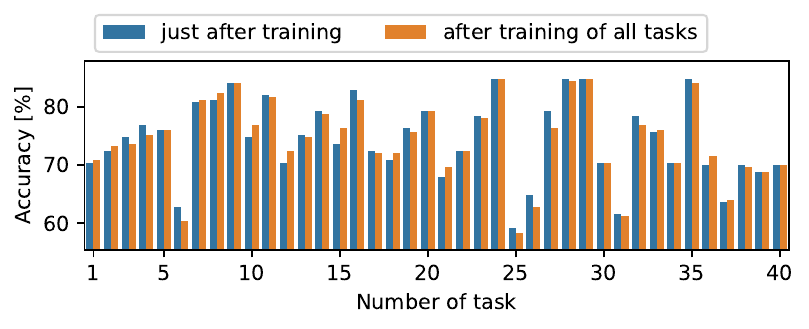}%
            \includegraphics[trim={0,2cm, 1,1cm, 0,2cm, 0,2cm},clip,width=0.5\linewidth]{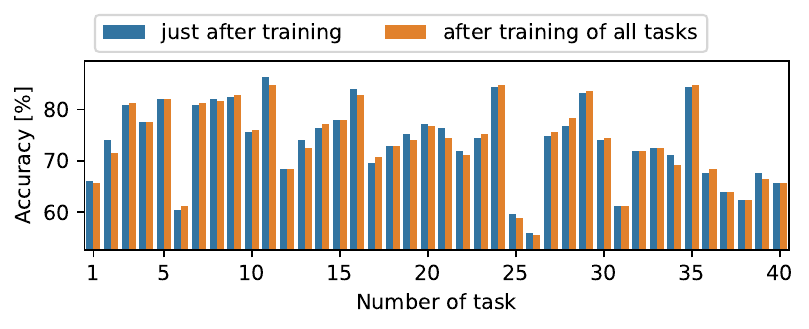}
            \includegraphics[trim={0,2cm, 1,1cm, 0,2cm, 0,9cm},clip,width=0.5\linewidth]{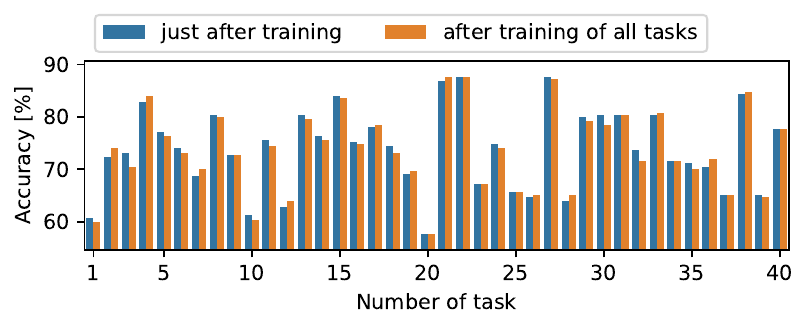}%
            \includegraphics[trim={0,2cm, 1,1cm, 0,2cm, 0,9cm},clip,width=0.5\linewidth]{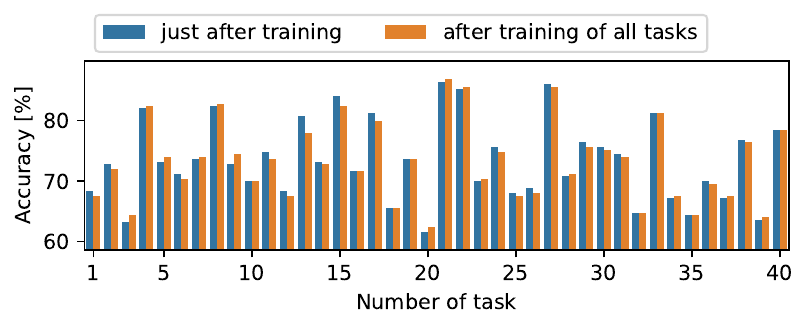}
            \includegraphics[trim={0,2cm, 1,1cm, 0,2cm, 0,9cm},clip,width=0.5\linewidth]{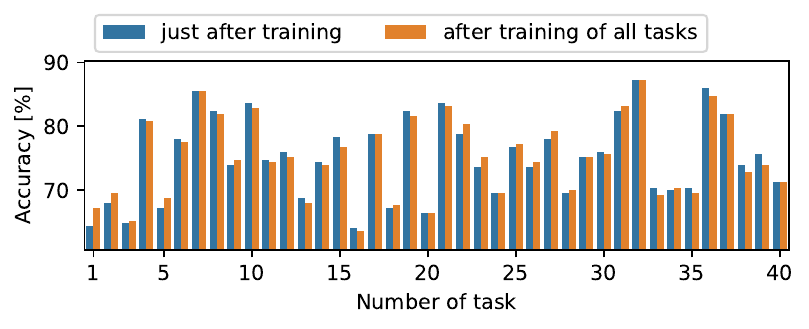}%
            \includegraphics[trim={0,2cm, 1,1cm, 0,2cm, 0,9cm},clip,width=0.5\linewidth]{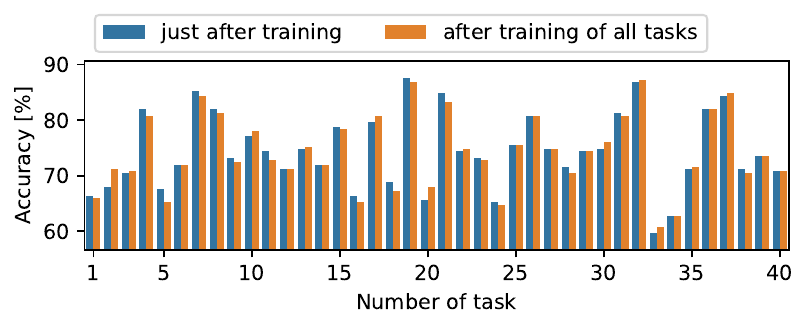}
            \includegraphics[trim={0,2cm, 1,1cm, 0,2cm, 0,9cm},clip,width=0.5\linewidth]{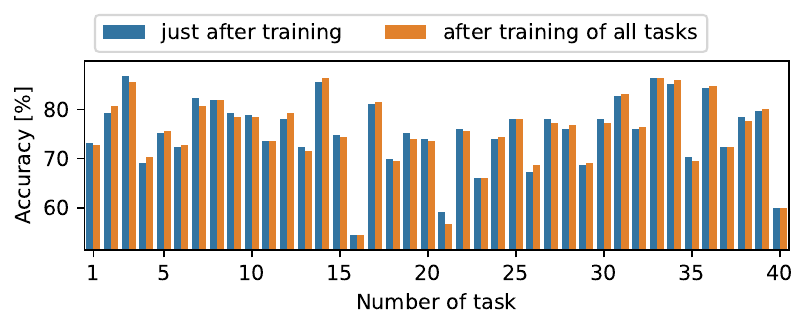}%
            \includegraphics[trim={0,2cm, 1,1cm, 0,2cm, 0,9cm},clip,width=0.5\linewidth]{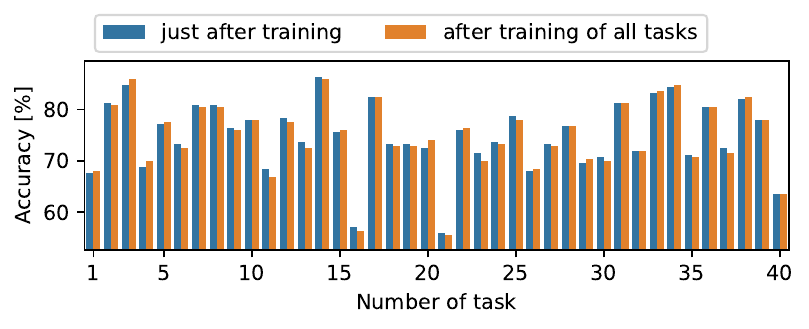}
            \includegraphics[trim={0,2cm, 0,2cm, 0,2cm, 0,9cm},clip,width=0.5\linewidth]{img/accuracy_fixed_model_4_Tiny_ImageNet_ResNet-18.pdf}%
            \includegraphics[trim={0,2cm, 0,2cm, 0,2cm, 0,9cm},clip,width=0.5\linewidth]{img/accuracy_trainable_model_4_Tiny_ImageNet_ResNet-18.pdf}
       \end{subfigure}
    \caption{Visualization of the overall accuracy of \our{} for all ten models trained through 40 CL tasks with samples from the Tiny ImageNet dataset. The left column presents scores for HyperMask with the fixed ResNet-18 target architecture that finally achieved $74.30$, $73.70$, $75.13$, $75.31$ and $76.04\%$ of mean task accuracy, respectively. HyperMask with the trainable ResNet-18 architecture, shown in the second column, was slightly less performing ($73.43$, $73.21$, $74.30$, $74.67$ and $75.54\%$ of mean task accuracy, respectively). Because the order of classes was randomly selected for each training run, it was impossible to present one graph for five models of a given architecture (with mean and confidence intervals defined per task). The models worked on differently constructed tasks. However, models from a given row may be compared to each other because the task and class settings were the same in such a case.\label{tinyimagenet:full_accuracy}}
\end{figure*}

\subsection{Analysis of consecutive CL tasks}
In the main work, we selected Split CIFAR-100 to visualize performance drop after learning subsequent CL tasks by two \our{} variants, see Fig.~\ref{cifar_100:accuracy_matrix}. The other perspective for this scenario gives Fig.~\ref{cifar:accuracy)tasks}. We can conclude that classification accuracy is stable for fixed target network models and only slightly changes with learning the next tasks. Trainable target networks gradually forget previous knowledge, although mean results are still competitive. Perhaps, in the future, one should consider other target network regularizers for convolutional neural networks, including different values of learning rates for hyper- and target networks in \our{}. 

For one of the easiest scenarios, i.e. Split MNIST, trainable \our{}, similarly as HNET, mostly retain information from the previous tasks, compare Fig.~\ref{split:accuracy_matrix}. The accuracy matrices for Permuted MNIST are presented in Fig.~\ref{permuted_10:accuracy_matrix}. Both \our{} and HNET feature minimal forgetting after training of consecutive CL tasks. However, \our{} with a trainable target model preserves high accuracy for the first task and performs well on the last learned tasks while HNET achieves quite a low performance for the newest CL tasks, keeping knowledge from the first ones. When we analyze the Permuted MNIST scenario with 100 tasks, we can conclude that \our{} strongly remembers the first task and its accuracy is higher than many subsequent CL tasks, see. Fig.~\ref{permuted:best_set_10_tasks}. It may be related to learning of both networks simultaneously and higher significance of the first task. We have to emphasize that all experiments with the MNIST dataset were performed with multilayer perceptrons as target models.

For another dataset, for which convolutional neural networks were selected as target networks, i.e. Tiny ImageNet, knowledge forgetting was
slight, both for fixed and trainable variants of the target network. Despite 40 CL tasks, knowledge was mainly preserved which may also be observed through low values of backward transfer presented in Table~\ref{tab:transfer}. For some tasks, a slight increase in classification accuracy was denoted after finishing the full training pipeline, see Fig.~\ref{tinyimagenet:full_accuracy}.  

\section{Appendix: Architecture details}\label{app:architectures}
We implemented \our{} in Python 3.7.16 with the use of such libraries like hypnettorch 0.0.4~\cite{von2019continual}, PyTorch 1.5.0, NumPy 1.21.6, Pandas 1.3.5, Matplotlib 3.5.3, seaborn 0.12.2 and others. Network training sessions were performed using NVIDIA GeForce RTX 2070, 2080 Ti, 3080 and 4090 graphic cards.

We tried to implement hypernetwork and target network architectures close to the work presenting HNET algorithm~\cite{von2019continual}, but we performed an intensive grid search optimization for some hyperparameters, especially those present only in \our{}. In all cases, we did not use chunked hypernetworks, i.e. we did not generate a mask in small pieces. It means that the hypernetwork output layer always had the same number of neurons, like the number of weights of the target network. This is because each output neuron produces a single value of the mask for the corresponding target network's weight. This solution is more memory expensive than the chunking approach, but, currently, in the case of \our{}, it ensures higher classification accuracy.

\paragraph{Permuted MNIST} Final experiments on the Permuted MNIST dataset with 10 CL tasks have been performed using the following architecture. The hypernetwork had two hidden layers with 100 neurons per each. The target network was selected as a multilayer perceptron with two hidden layers of 1000 neurons and ELU activation function with $\alpha$ hyperparameter regarding the strength of the negative output equaling 1. The size of the embedding vectors was set to 24. The sparsity parameter $p$ was adjusted to 0, and the regularization hyperparameters were as follows: $\beta = 0.0005$ and $\lambda = 0.001$. Furthermore, a masked $L_1$ regularization was chosen. The training of models was performed through 5000 iterations with a batch size of $128$ and Adam optimizer with a learning rate set to $0.001$. Finally, models after the last training iterations were selected. The validation set consisted of 5000 samples. The data was not augmented. The presented results were averaged over five training runs for different seed values. Also, the dataset was padded with zeros, therefore the MNIST images' final size was $32 \times 32$.

For 100 CL tasks, the hyperparameters were the same as above, but three training runs were performed.

To select the best hyperparameter set, we performed an intensive hyperparameter optimization. In the final stage, we evaluated, in different configurations, various hypernetwork settings ($[25, 25], [100, 100]$), masked and non-masked $L_1$ regularization, $p \in \lbrace 0, 30 \rbrace$, $\beta \in \lbrace 0.0005, 0.001, 0.0025, 0.005, 0.01, 0.1 \rbrace$ and $\lambda \in \lbrace 0.0005, 0.001, 0.0025, 0.005, 0.01 \rbrace$. 

In the initial experiments, we also considered hypernetworks having two hidden layers with 50 neurons per each, embeddings of sizes 8 and 72, a learning rate of $0.0001$, batch size of $64$, $\beta = 0.05$, $\lambda \in \lbrace 0.0001, 0.00001, 0.05 \rbrace$ and $p = 70$.

In the case of the fixed target network scenario, we changed the $\beta$ regularization hyperparameter to $0.0001$ because the model performance was higher than for $\beta = 0.0005$. Still, the lack of network sparsification was a better option. 

\paragraph{Split MNIST} For this dataset with 5 CL tasks, we applied data augmentation and trained models through 2000 iterations. The best-performing model comprised a hypernetwork with two hidden layers with 25 neurons per each and a target network with two hidden layers consisting 400 neurons. We used $\beta = 0.001$ and a sparsity parameter $p=30$. Furthermore, the embedding size was 128. In each task, 1000 samples were assigned to the validation set. The rest of the hyperparameters were the same as for Permuted MNIST, i.e. we applied a masked $L_1$ regularization with $\lambda = 0.001$, ELU activation function with $\alpha = 1$, Adam optimizer with a learning rate of $0.001$ and a batch size of $128$. Also, the mean results were averaged over five training runs.

During the hyperparameter optimization stage, we evaluated models with embedding sizes of $24, 72, 96$ and $128$, hypernetworks with hidden layers of shapes $[10, 10], [25, 25]$ and $[50, 50]$, masked and non-masked $L_1$ regularization, batch sizes of $64$ and $128$, $\beta \in \lbrace 0.001, 0.01 \rbrace$, $p \in \lbrace 0, 30, 70 \rbrace$ and $\lambda \in \lbrace 0.0001, 0.001 \rbrace$.

When the target network was fixed during training, we used similar hyperparameters as in the trainable target scenario, but we changed $\beta$ to 0.0001. We also verified results for $\beta \in \lbrace 0.0001, 0.001, 0.01, 0.1 \rbrace$ and sparsity parameter $p \in \lbrace 0, 30 \rbrace$.

\paragraph{CIFAR-100}
\subparagraph{Known task identity}
In this dataset, we assumed 10 tasks with ten classes per each (classes 1-10, 11-20, ..., 91-100). Another version of this CL benchmark adopts CIFAR-10 and 5 tasks (i.e., 50 classes) of the CIFAR-100 dataset, like in~\cite{von2019continual}. However, we selected the first scenario, similarly as in~\cite{kang2022forget}. 

We performed experiments with fixed and trainable target architectures. As the target network, we used a well-known ResNet-18 architecture~\cite{he2015ResNet} consisting of five groups of convolutional layers. The initial group has a single convolutional layer with 16 feature maps and a single max pooling layer. The following four groups have two blocks of two convolutional layers, with 16, 32, 64 and 128 feature maps, respectively. Thus, in total, we have 17 convolutional and a single fully-connected layer. Also, skip connections are realized through one-dimensional convolutions when shapes are changed.


To ensure a fair comparison with WSN~\cite{kang2022forget}, we trained \our{} through 200 epochs with batch size 64. Similarly, the initial learning rate was $10^{-3}$ and was multiplied by $0.5$ (until $10^{-6}$) if no improvement was noted through 6 consecutive epochs. The same setup was used in WSN. According to the remaining hyperparameters, we selected an embedding size of 48, a hypernetwork of two hidden layers with 100 neurons, and the best model was selected using the minimum validation loss value criterion. When the target network was fixed through the training process, as $\beta$ we selected $0.1$ and the sparsity parameter $p$ was set to 0 (pruning was not performed). When the target model was trainable, we additionally chose $1$ as the $\lambda$ value, and we used a non-masked $L_1$ regularization. Also, the sparsity parameter $p$ was changed to $15\%$. In both cases, batch normalization was used (these layers were excluded from multiplying by hypernetwork-based masks). Batch statistics were calculated even during the evaluation, i.e. parameters were not stored after consecutive CL tasks. 

For both considered networks, the dataset was augmented according to the approach implemented in the hypnettorch library. For each task, the validation set consisted of 500 samples, 50 per class. It is worth emphasizing that the lack of data augmentation led to significantly lower accuracy than in the opposite case.

For WSN, we followed the author's choice of $5\%$ training data to the validation set and the sparsity parameter $c=0.5$ as it was the highest-performing option for AlexNet.

\subparagraph{Unknown task identity}
In this scenario, we followed the setting from~\cite{goswami2023fecam} to ensure a representative comparison of methods. The order of classes present in CIFAR-100 was mixed, and we followed the order described in the FeCAM source code. Then, 100 classes were divided into five tasks, with 20 classes per each. This class incremental setting corresponds to the results from Table 5 depicted in~\cite{goswami2023fecam}. We also followed the accuracy calculation approach from this work, which is different than in the previously described experiments in our paper. Therefore, average task accuracy presented in Table~\ref{tab:CL3} in Section~\ref{experiments} is calculated as follows:
\begin{equation*}
    \mathrm{AVG} = \frac{1}{5} \sum\limits_{j=1}^{5} \mathrm{Acc_j},
\end{equation*}
\noindent where $\mathrm{Acc_j}$ represents the overall accuracy for the test samples from tasks $\lbrace 1, 2, ..., j \rbrace$ after creating class prototypes from the $j$--th task. The last task accuracy just refers to $\mathrm{Acc_5}$, which is calculated for the test set containing samples from all 100 classes after completing the fifth task.

Furthermore, we present results for the best-trained model as well as for 5 different runs with the same hyperparameters and the given order of CIFAR-100 classes.

In this case, we used the same hyperparameters for \our{} as for known task identity scenarios. However, the augmentation of the dataset was performed differently. In this setup, we followed the more sophisticated approach from FeCAM.

\paragraph{Tiny ImageNet}
In this case, we divided the dataset randomly into 40 tasks with five classes, similarly to~\cite{kang2022forget}. We assumed the same training strategy, i.e. we learned each task through 10 epochs, and the validation set consisted of 250 samples. We performed experiments with the ResNet-18 architecture, explained in detail in the section devoted to CIFAR-100. However, in WSN~\cite{kang2022forget}, La-MaML~\cite{gupta2020lamaml} and FS-DPGM~\cite{deng2021flattening} authors used an architecture with four convolutional and three fully connected layers. Therefore, we additionally calculated experiments with ResNet-18 for one of our main baselines, i.e. WSN. 

The best models were chosen according to the values of the validation loss. We augmented the dataset using random cropping and horizontal flipping. For the hypernetwork, we selected a multilayer perceptron with two hidden layers of 200 neurons while the task embedding vector size was set to 256. We performed training with Adam optimizer with a batch size of 16 and the initial value of the learning rate $10^{-4}$. The learning rate scheduler was the same as for CIFAR-100, i.e. a patience step was equal to 6 epochs, and the multiplication factor was $0.5$. When the target model was fixed, the $\beta$ regularization parameter was equal to $1$ while the sparsity parameter $p$ was at the level of $70\%$. In the opposite case, we used non-masked $L_1$ regularization with $\beta = 10$ and $\lambda = 0.1$. Also, we selected the sparsity parameter $p$ as $15 \%$.

In the hyperparameter optimization stage, we considered different sizes of the embedding vectors, i.e. $48, 96, 128, 256$, various structures of the hypernetwork (with a single hidden layer having 100 or 200 neurons or two hidden layers having 200 neurons). We evaluated different values of the $\beta$ regularization hyperparameter, i.e. $0.01, 0.1, 1, 10$ or even $100$, and for the trainable target model $\lambda$ regularization values among $0.01, 0.1, 1, 10$. We selected the sparsity parameter from $0, 15, 30, 50, 70$ and even $90\%$ as candidates. We also experimented with different learning rates, i.e. $10^{-4}, 10^{-3}$ and $5 \cdot 10^{-3}$.

For WSN~\cite{kang2022forget}, we selected the same number of epochs, batch size, learning rate and scheduler strategy as in \our{}. Following the authors' source code setting, we set $10\%$ as a fraction of the training data used for the validation set. Also, we selected $0.5$ as the sparsity parameter $c$.
\section{Appendix: Influence of semi-binary mask on classification task in trainable MLPs}
\begin{figure}[H]
        \centering
                \includegraphics[trim={1cm, 0,7cm, 0,2cm, 0,2cm},clip,width=0.35\linewidth]{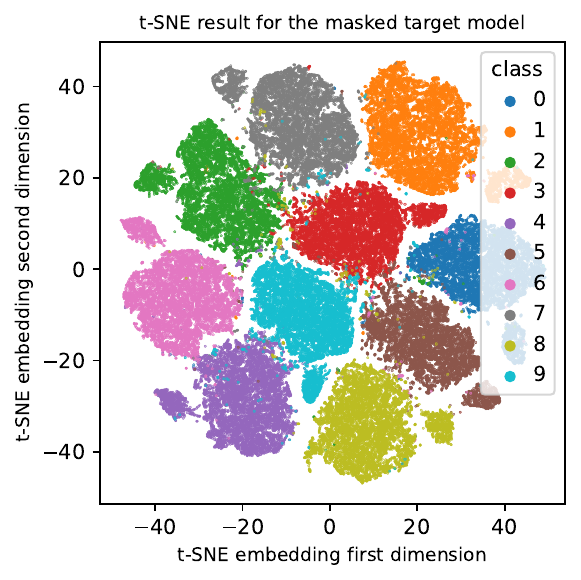}%
                \includegraphics[trim={0,6cm, 0,7cm, 0,2cm, 0,2cm},clip,width=0.405\linewidth]{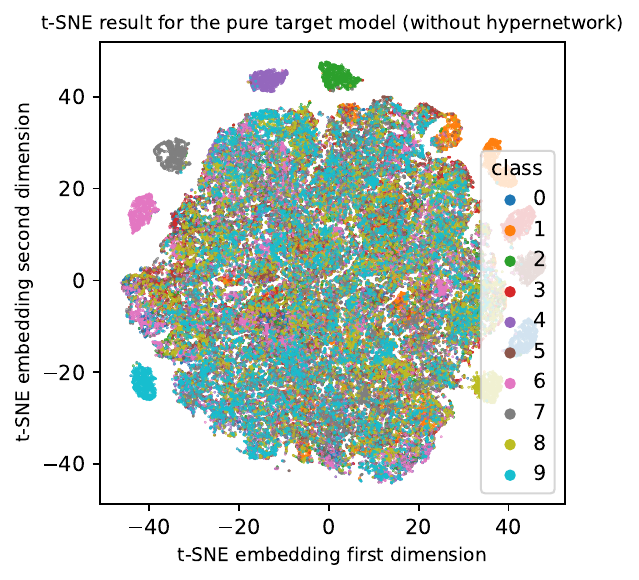}
        \caption{Visualization of t-SNE embeddings based on a target network's output classification layer activations for Permuted MNIST. The left-hand side plot corresponds to the trainable target network weighted by a semi-binary mask (\our{}+T), while the plot on the right refers just to the trainable target network, without a mask produced by the hypernetwork. In the first case, data sample classes are separated; in the second case, only samples from the first task are distinguished. \label{permuted:tSNE}}
    \end{figure}

\begin{figure}[!ht]
        \centering
            \includegraphics[trim={0,2cm, 0,2cm, 0,0cm, 0,2cm},clip,width=0.38\linewidth]{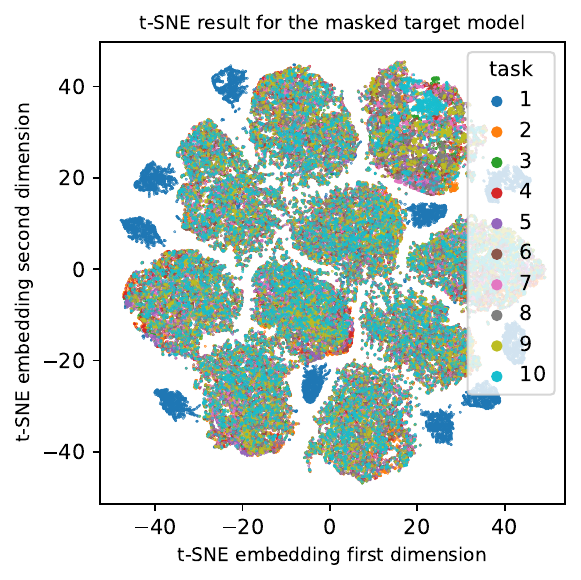}%
            \includegraphics[trim={0,0cm, 0,2cm, 0,1cm, 0,2cm},clip,width=0.42\linewidth]{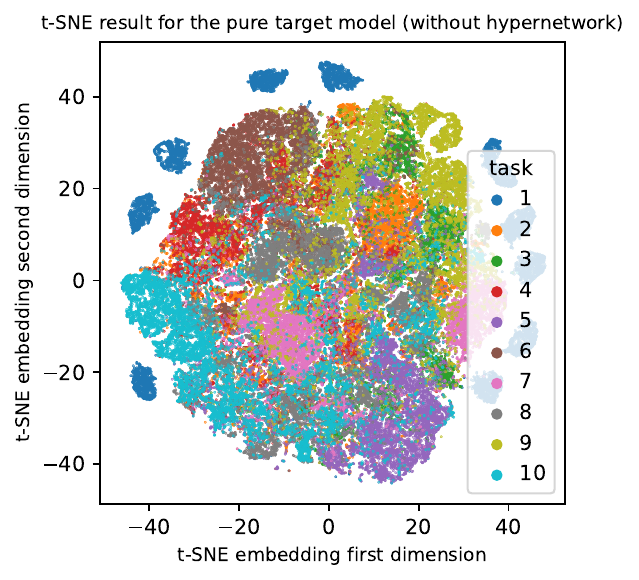}
        \caption{t-SNE embeddings of features extracted from all data samples of 10 tasks of Permuted MNIST dataset, created similarly as embeddings presented in Fig.~\ref{permuted:tSNE}, but the samples are labelled relative to the CL tasks. On the left column, results for \our{}+T (i.e. hypernetwork and target network) are shown. On the right column are presented only results for the target network, without the application of a mask from hypernetwork. The plots indicate that samples from the first task form a separate structure in the data space.\label{permuted:tSNE_task}}
    \end{figure}

\begin{figure}[!ht]
        \centering
                \includegraphics[trim={0,2cm, 0,2cm, 1,35cm, 0,2cm},clip,width=0.38\linewidth]{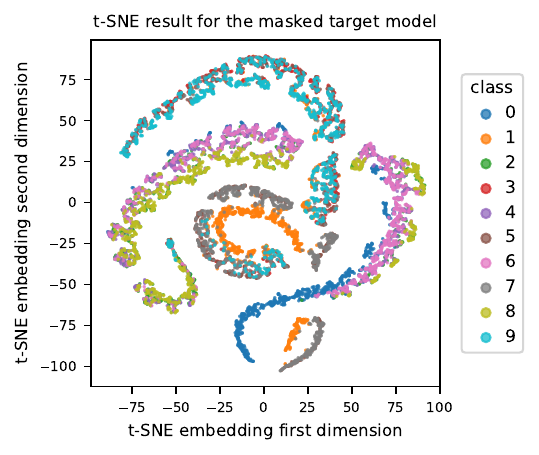}%
                \includegraphics[trim={0,0cm, 0,2cm, 0,1cm, 0,2cm},clip,width=0.46\linewidth]{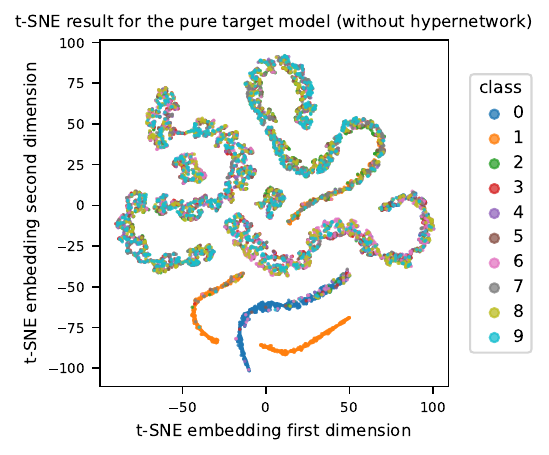}
                \includegraphics[trim={0,2cm, 0,2cm, 1,35cm, 0,2cm},clip,width=0.38\linewidth]{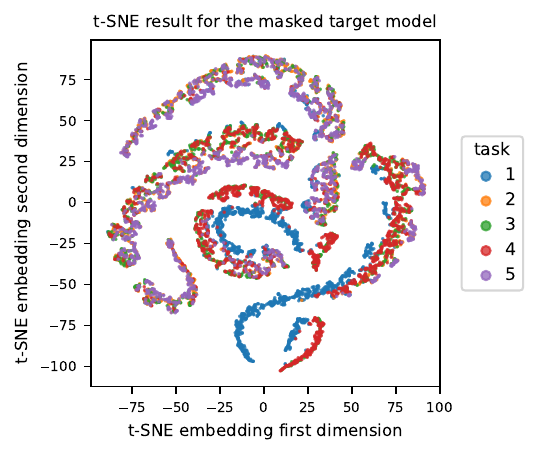}%
                \includegraphics[trim={0,0cm, 0,2cm, 0,1cm, 0,2cm},clip,width=0.46\linewidth]{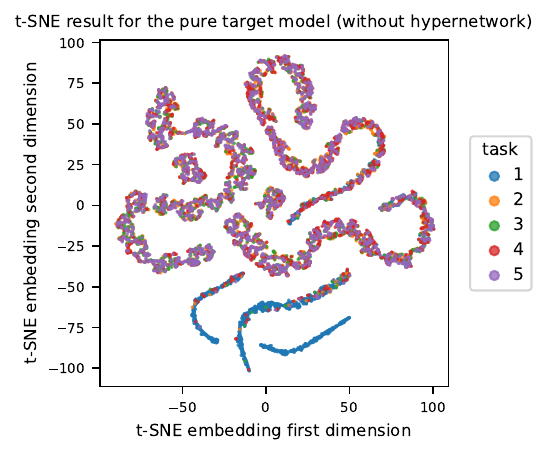}
        \caption{t-SNE embeddings of features extracted from all data samples of 5 tasks of the Split MNIST dataset. Values were taken from the classification layer of the target network for an exemplary model that achieved $99.76\%$ overall accuracy after 5 CL tasks. The results for a tandem hypernetwork and target network (like in \our+T) are presented in the left column. On the right column is shown a mask from hypernetwork that was not applied to the target network while it was trained in tandem in \our{}+T. In the first case, classes form different clusters, especially the pairs of classes which were mutually compared in consecutive CL tasks (0 and 1, 2 and 3, etc.). In the second case, only 0s and 1s are separated while the remaining data samples are mixed in the embedding subspace. Furthermore, in this situation, data from the first task form a separate cluster, which suggests that it mainly defines the structure of the data space. \label{split:tSNE}}
    \end{figure}

The semi-binary mask of \our{} helped the trainable target network to discriminate classes in consecutive CL tasks, when a target model was multilayer perceptron.  We took the fully-trained model on Permuted or Split MNIST datasets and collected activations of the output classification layer of the target network. In Fig.~\ref{permuted:tSNE}, we present t-SNE two-dimensional embeddings obtained from the classification layer output for all data samples from 10 tasks, separated by a class identity, while in Fig.~\ref{permuted:tSNE_task} we show the samples divided according to the task number. Interestingly, even when the mask was not applied, the first task was solved correctly, and the corresponding samples formed separate clusters. The evaluation was performed for an exemplary model that achieved $97.72\%$ of mean overall accuracy after 10 CL tasks. The results for a tandem hypernetwork and target network (like in \our+T) are presented on the left side. On the right side is shown a situation in which a mask from the hypernetwork was not applied to the target network trained in \our{}+T. In the first case, data sample classes are clearly separated; in the second case, only samples from the first task are distinguished. The remaining data samples form one cluster in the embedding space. An analogous plot for Split MNIST is depicted in Fig.~\ref{split:tSNE}. The first task samples form a different structure. However, the division is more ambiguous than for Permuted MNIST.


\section{Appendix: Importance of the trainable MLP's target network in HyperMask}
\label{app:target_analysis}

Based on the performed experiments we can conclude that the application of trainable target networks may be beneficial for multilayer perceptrons. Therefore, we evaluated changes in the target weights after consecutive CL tasks for a selected model trained on Permuted MNIST with hyperparameters described in Appendix~\ref{app:architectures}. Tables~\ref{tab:targetDistance1}--\ref{tab:targetDistance3} present distances between the target weights of a given network layer, calculated as follows: 

\begin{equation*}
    D_{i,j}^{l_k} = \sum\limits_{w \in l_k} |w_i - w_j|,
\end{equation*}

\noindent where $i$ and $j$ denote task identities, $k \in \lbrace 1, 2, 3 \rbrace$ corresponds to the number of the network layer, $l_k$ denotes the set of the $k$--th layer weights, and $w_i$ is a value of the weight $w$ after learning of the $i$--th task.

We can observe that the sum of the absolute differences between consecutive target weights after subsequent CL tasks is strictly increasing. The values of the difference $D_{1, 10}^{l_k} - D_{1, 2}^{l_k}$ are equal to $695$, $302.75$ and $12.22$ for the layers $l_1$, $l_2$, $l_3$, having $1.025.000$, $1.001.000$ and $10.010$ neurons, respectively. This proves the plasticity of the target, despite the non-forgetting of previous tasks' knowledge, see the matrix on the left-hand side in Fig.~\ref{permuted_10:accuracy_matrix}.
\ctable[
caption = {Distances between target network's weights of the first hidden layer after training of consecutive CL tasks for an exemplary model trained on the Permuted MNIST dataset with hyperparameters described in Appendix~\ref{app:architectures}. The position $(i, j)$ denotes a distance between the network's weights after the $i$--th task and the $j$--th task. The first hidden layer contains $1000 \times 1024$ neuron connections and $1000$ bias values ($1.025.000$ individual values). Since the matrix is symmetric, only distances above the main diagonal are given.},
label = tab:targetDistance1,
mincapwidth = \textwidth,
doinside = {\footnotesize},
star = False,
pos = h
]{rrrrrrrrrrr}{}{\FL
    \textbf{task} & \textbf{1} & \textbf{2} & \textbf{3} & \textbf{4} & \textbf{5} & \textbf{6} & \textbf{7} & \textbf{8} & \textbf{9} & \textbf{10} \ML
    \textbf{1} & & 148.01 & 293.95 & 406.84 & 498.70 & 569.80 & 648.47 & 720.51 & 785.53 & 843.01 \\
    \textbf{2} & & & 186.94 & 320.83 & 425.43 & 504.50 & 588.97 & 665.50 & 734.11 & 794.08 \\ 
    \textbf{3} & & & & 181.17 & 310.16 & 404.57 & 500.48 & 584.35 & 658.86 & 723.21 \\
    \textbf{4} & & & & & 176.20 & 292.82 & 405.71 & 500.66 & 582.78 & 652.29 \\ 
    \textbf{5} & & & & & & 161.76 & 300.25 & 410.62 & 502.33 & 578.91 \\
    \textbf{6} & & & & & & & 183.81 & 318.21 & 423.98 & 508.82 \\
    \textbf{7} & & & & & & & & 180.29 & 311.77 & 412.42 \\ 
    \textbf{8} & & & & & & & & & 175.31 & 301.64 \\ 
    \textbf{9} & & & & & & & & & & 169.94 \\ 
    \textbf{10} & & & & & & & & & & \LL
}

\ctable[
caption = {Distances between target network's weights of the second hidden layer after training of consecutive CL tasks for an exemplary model trained on the Permuted MNIST dataset with hyperparameters described in Appendix~\ref{app:architectures}. The position $(i, j)$ denotes a distance between the network's weights after the $i$--th task and the $j$--th task. The second hidden layer contains $1000 \times 1000$ neuron connections and $1000$ bias values ($1.001.000$ individual values). Since the matrix is symmetric, only distances above the main diagonal are given.},
label = tab:targetDistance2,
mincapwidth = \textwidth,
doinside = {\footnotesize},
star = False,
pos = h
]{rrrrrrrrrrr}{}{\FL
    \textbf{task} & \textbf{1} & \textbf{2} & \textbf{3} & \textbf{4} & \textbf{5} & \textbf{6} & \textbf{7} & \textbf{8} & \textbf{9} & \textbf{10} \ML
    \textbf{1} & & 94.17 & 167.06 & 219.27 & 256.19 & 284.35 & 322.32 & 350.43 & 375.09 & 396.92 \\
    \textbf{2} & & & 124.70 & 187.48 & 229.56 & 260.77 & 301.23 & 331.00 & 356.91 & 379.74 \\ 
    \textbf{3} & & & & 124.41 & 180.41 & 218.27 & 264.40 & 297.60 & 326.07 & 350.76 \\
    \textbf{4} & & & & & 115.52 & 166.84 & 222.46 & 260.75 & 292.62 & 319.84 \\ 
    \textbf{5} & & & & & & 106.55 & 178.28 & 223.87 & 260.08 & 290.37 \\
    \textbf{6} & & & & & & & 129.12 & 186.69 & 228.78 & 262.71 \\
    \textbf{7} & & & & & & & & 117.57 & 174.29 & 215.87 \\ 
    \textbf{8} & & & & & & & & & 113.90 & 169.12 \\ 
    \textbf{9} & & & & & & & & & & 110.84 \\ 
    \textbf{10} & & & & & & & & & & \LL
}

\ctable[
caption = {Distances between target network's weights of the output classification layer after training of consecutive CL tasks for an exemplary model trained on the Permuted MNIST dataset with hyperparameters described in Appendix~\ref{app:architectures}. The position $(i, j)$ denotes a distance between the network's weights after the $i$--th task and the $j$--th task. The output classification layer contains $1000 \times 10$ neuron connections and $10$ bias values ($10.010$ individual values). Since the matrix is symmetric, only distances above the main diagonal are given.},
label = tab:targetDistance3,
mincapwidth = \textwidth,
doinside = {\footnotesize},
star = False,
pos = h
]{rrrrrrrrrrr}{}{\FL
    \textbf{task} & \textbf{1} & \textbf{2} & \textbf{3} & \textbf{4} & \textbf{5} & \textbf{6} & \textbf{7} & \textbf{8} & \textbf{9} & \textbf{10} \ML
    \textbf{1} & & 4.57 & 7.23 & 9.10 & 10.92 & 12.12 & 12.90 & 14.37 & 15.82 & 16.79 \\
    \textbf{2} & & & 5.04 & 7.28 & 9.23 & 10.69 & 11.80 & 13.33 & 14.76 & 15.81 \\ 
    \textbf{3} & & & & 4.50 & 6.66 & 8.10 & 9.56 & 11.15 & 12.65 & 13.92 \\
    \textbf{4} & & & & & 4.23 & 6.31 & 7.95 & 9.79 & 11.56 & 12.73 \\ 
    \textbf{5} & & & & & & 3.83 & 6.34 & 8.42 & 10.15 & 11.65 \\
    \textbf{6} & & & & & & & 4.43 & 7.27 & 9.13 & 10.88 \\
    \textbf{7} & & & & & & & & 4.70 & 7.33 & 8.86 \\ 
    \textbf{8} & & & & & & & & & 4.81 & 6.85 \\ 
    \textbf{9} & & & & & & & & & & 3.99 \\ 
    \textbf{10} & & & & & & & & & & \LL
}

\section{Appendix: Classification with task identity recognition}\label{app:task_recognition}
\subsection{\our{} + ENT}
\our{} + ENT is a simple method where task recognition is based on a minimization of an entropy criterion calculated on network logits' outputs. A similar approach has been used in HNET~\cite{von2019continual} for strategies in which task identity was identified without the generation of additional data samples. The consecutive steps of this method are summarized in Algorithm~\ref{alg:hypermask+ent}.

Let us assume that we have a hypernetwork $\mathcal{H}$ with weights $\bm{\Theta}$ and a target network $f$ with weights $\bm{\theta}$, trained according to Algorithm~\ref{alg:hypermask}. Depending on the considered scenario, we can evaluate the method after finishing the training of all $T$ tasks or just after one of the previous tasks, $t$, where $t \in \lbrace 1, 2, ..., T \rbrace$. In the second case, we only predict the samples that may belong to one of the classes $\lbrace 1, 2, ..., t \rbrace$.

In the general case, we suppose that \our{} was trained on $t$ tasks, and we have to classify a test sample $\bm{x}$. Using embeddings designed for all possible tasks, we perform forward propagation through the hyper- and target network to get logits of the output classification layer. In the next step, we apply the softmax function to obtain a vector of predictions in which single values range from 0 to 1, according to the following equation:
\begin{equation*}
\bm{\psi}_{i}(\bm{x}) = \mathrm{Softmax}\big(f\big(\bm{x}, \mathcal{H}(\bm{e}_i, \Phi) \big) \big),
\end{equation*}
\noindent where $\bm{e}_i$ is the embedding of the $i$--th task. Therefore, we obtain $t$ vectors of \our{} predictions: $\big(\bm{\psi}_{1}(\bm{x}), \bm{\psi}_{2}(\bm{x}), ..., \bm{\psi}_{t}(\bm{x}) \big)$. Then, for such vectors, we calculate the entropy criterion:
\begin{equation*}
\psi_{i}^{ENT}(\bm{x}) = - \sum\nolimits_{j=1}^{C} \big( \psi_i(\bm{x})[j] \cdot \log \big( \psi_i(\bm{x})[j] \big) \big),
\end{equation*}
\noindent where $C$ is the number of classes. Now, we have values corresponding to each of $t$ tasks: $\big( \psi_{1}^{ENT}(\bm{x}), \psi_{2}^{ENT}(\bm{x}), ..., \psi_{t}^{ENT}(\bm{x}) \big)$. We select a task minimizing the entropy criterion:
\begin{equation*}
t^{\star} = \argmin\limits_{i=1,2,...,t} \big( \psi_{i}^{ENT}(\bm{x}) \big).
\end{equation*}
Finally, we choose the class with the highest softmax value and assign it as the label for sample $x$:
\begin{equation*}
c^{\star} = \argmax\limits_{j=1,2,...,C} \big( \psi_{t^{\star}}(\bm{x})[j] \big).
\end{equation*}

\begin{algorithm}[!t]
    \caption{The pseudocode of \our{}+ENT.\label{alg:hypermask+ent}}
    \begin{algorithmic}
    \Require trained hypernetwork $\mathcal{H}$ with weights $\bm{\Phi}$, including embeddings $(\bm{e}_1, \bm{e}_2, ..., \bm{e}_T)$; trained target network $f$ with weights $\bm{\theta}$; test samples $\bm{X}_{test}^{t}$; selected task $t$, where $t \in \lbrace 1, 2, ..., T \rbrace$; number of classes $C$
    \Ensure predictions for test samples $\bm{\hat{Y}}_{test}^t$
    \vspace{0,1cm}
    \For{$\bm{x} \in \bm{X}_{test}^t$}
        \For{$i \in \lbrace 1, 2, ..., t \rbrace$}
            \State $\bm{\psi}_{i}(\bm{x}) = \mathrm{Softmax}\big(f\big(\bm{x}, \mathcal{H}(\bm{e}_i, \Phi) \big) \big)$ 
            \State $\psi_{i}^{ENT}(\bm{x}) = - \sum\nolimits_{j=1}^{C} \big( \psi_i(\bm{x})[j] \cdot \log \big( \psi_i(\bm{x})[j] \big) \big)$
        \EndFor
            \State $t^{\star} = \argmin\limits_{i=1,2,...,t} \big( \psi_{i}^{ENT}(\bm{x}) \big)$
            \State $c^{\star} = \argmax\limits_{j=1,2,...,C} \big( \psi_{t^{\star}}(\bm{x})[j] \big)$
    \EndFor
    \end{algorithmic}
    \end{algorithm}

\subsection{\our{} + FeCAM}
\our{} + FeCAM is a hybrid method which combines the advantages of these two approaches. \our{} is trained similarly to the basic case, but then it is used as the feature extractor while FeCAM assigns a label based on the \our{} output. Initially, in FeCAM~\cite{goswami2023fecam}, the feature extractor was not trained after the first task. We propose this strategy for cases where task identities are unknown. The pseudocode of this method is depicted in Algorithm~\ref{alg:hypermask+fecam} and now will be described in detail.

Similar to \our{} + ENT, let us suppose that we have a hypernetwork $\mathcal{H}$ with weights $\bm{\Theta}$ and a target network $f$ with weights $\bm{\theta}$, trained according to Algorithm~\ref{alg:hypermask}. During the evaluation phase of the $t$--th task, where $t \in \lbrace 1, 2, ..., T \rbrace$ and $T$ is the total number of tasks, we have to prepare class prototypes based on training samples belonging to consecutive classes present in all previous tasks, up to the $t$-th task. In practice, we can just update the set of prototypes in subsequent tasks. There is no need to recalculate prototypes from the preceding tasks if no new samples from previously known classes exist. 

In the beginning, for each element of the training dataset, we extract the output of the hidden layers of the target network after forward propagation (just before the last fully connected layer). It means that for the samples from the $t$--th task, we prepare features extraction:
\begin{equation*}
\bm{\tilde{X}}^t = f\big(\bm{X}_{train}^t, \mathcal{H}(\bm{e}_t, \Phi) \big).
\end{equation*}

Based on such prepared features, for each class from the training set, we calculate its prototype as an average of all samples belonging to this class:
\begin{equation*}
\bm{\mu}_c = \frac{1}{|\bm{X}_c|} \sum\limits_{\bm{x} \in \bm{X}_c}\bm{x},
\end{equation*}
\noindent where $\bm{X}_c$ represents features calculated for all training samples from the $c$--th class while $|\bm{X}_c|$ is the number of these samples. We also have to calculate a covariance matrix, which has to be invertible. Similar to FeCAM~\cite{goswami2023fecam}, we apply a covariance shrinkage according to the following equation:
\begin{equation*}
(\bm{\Sigma}_c)_s = \bm{\Sigma}_c + D_1 \bm{I} + D_2 (\bm{1} - \bm{I}),
\end{equation*}
\noindent where $\bm{\Sigma}_c$ is the covariance matrix, $D_1$ is the mean value of all diagonal elements of $\bm{\Sigma}_c$, while $D_2$ is the mean value of the elements lying outside the main diagonal of $\bm{\Sigma}_c$. Also, $\bm{I}$ is the identity matrix, and $\bm{1}$ is the matrix filled with ones.

In \our{}+FeCAM, the covariance shrinkage is performed twice. Furthermore, contrary to FeCAM, we skip using Tukey's Ladder of Powers transformation due to worse performance. Finally, we normalize the covariance matrix:
\begin{equation*}
(\tilde{\bm{\Sigma}}_c)_s[i, j] = \dfrac{({\bm{\Sigma}}_c)_s [i, j]}{\sqrt{({\bm{\Sigma}}_c)_s [i, i]} \cdot \sqrt{({\bm{\Sigma}}_c)_s [j, j]}}, 
\end{equation*}
\noindent where $({\bm{\Sigma}}_c)_s [i, j]$ corresponds to the element of the shrunken covariance matrix located at position $[i, j]$.

When all class prototypes are determined, one can start evaluating the test set. Therefore, for a given test sample $\bm{x}$, it is necessary to collect the network output for each embedding $\bm{e}_i$, where $i \in \lbrace 1, 2, ..., t \rbrace$, and $t$ is the number of the last trained task:
\begin{equation*}
\bm{\phi}_{i}(\bm{x}) = f\big(\bm{x}, \mathcal{H}(\bm{e}_i, \Phi) \big).
\end{equation*}

As before, we use features from the last hidden layer of \our{}, not the output of the classification layer. Then, we must calculate the squared Mahalanobis distance between all classes and network outputs. Therefore, such a distance between features created by \our{} with the $i$--th embedding and the $c$--th class prototype is defined in the following way:
\begin{equation*}
d_M \big( \bm{\phi}_i(\bm{x}), \bm{\mu}_c \big) = \bigg( \frac{\bm{\phi}_i(\bm{x})}{\lVert \bm{\phi}_i(\bm{x}) \rVert_2} - \frac{\bm{\mu}_c}{\lVert \bm{\mu}_c \rVert_2} \bigg)^\top (\tilde{\bm{\Sigma}}_c)^{-1}_s \bigg( \frac{\bm{\phi}_i(\bm{x})}{\lVert \bm{\phi}_i(\bm{x}) \rVert_2} - \frac{\bm{\mu}_c}{\lVert \bm{\mu}_c \rVert_2} \bigg),
\end{equation*}
\noindent where $(\tilde{\bm{\Sigma}}_c)^{-1}_s$ is the inverse of a normalized covariance matrix. Vectors of features and class prototypes are also normalized. Finally, one needs just to choose the class whose prototype is the closest to the given sample representation:
\begin{equation*}
c^{\star} = \argmin\limits_{c=1,2,...,|\bm{\mu}|} \bigg( \min\limits_{i=1,2,...,t} d_M \left( \bm{\phi}_i(\bm{x}), \bm{\mu}_c \right) \bigg),
\end{equation*}
\noindent where $|\bm{\mu}|$ is the number of classes (and corresponding prototypes).

\begin{algorithm}[!t]
    \caption{The pseudocode of \our{}+FeCAM.\label{alg:hypermask+fecam}}
    \begin{algorithmic}
    	\Require trained hypernetwork $\mathcal{H}$ with weights $\bm{\Phi}$, including embeddings $(\bm{e}_1, \bm{e}_2, ..., \bm{e}_T)$; target network $f$ with weights $\bm{\theta}$; training datasets $ (D_{train}^{1}, D_{train}^2, ..., D_{train}^T)$, $  (\bm{X}_{train}^i, \bm{Y}_{train}^i) \in D_{train}^i,$ $i \in \lbrace 1, ..., T \rbrace$; test samples $(\bm{X}_{test}^{1}, \bm{X}_{test}^2, ..., \bm{X}_{test}^T)$
            \vspace{0,2cm}
    	\Ensure predictions for test samples $(\bm{\hat{Y}}_{test}^1, \bm{\hat{Y}}_{test}^2, ..., \bm{\hat{Y}}_{test}^T)$
    	\vspace{0,1cm}
        \For{$t \leftarrow 1$ to $T$}
            \For{$i \leftarrow 1$ to $t$}
                \State $\bm{\hat{X}}^i = f\big(\bm{X}_{train}^i, \mathcal{H}(\bm{e}_i, \Phi) \big)$
            \EndFor
            \State $\bm{\tilde{X}}^t = \bm{\hat{X}}^1 \cup \bm{\hat{X}}^2 \cup ... \cup \bm{\hat{X}}^t$
            \State $\bm{\tilde{Y}}^t = \bm{Y}_{train}^1 \cup \bm{Y}_{train}^2 \cup ... \cup \bm{Y}_{train}^t$
            \vspace{0,1cm}
            \For{$c \in \bm{\tilde{Y}}^t$}
                \State Select samples belonging to the $c$--th class, $\bm{X}_c$, from $\bm{\tilde{X}}^t$ 
                \State $\bm{\mu}_c = \frac{1}{|\bm{X}_c|} \sum\limits_{\bm{x} \in \bm{X}_c}\bm{x}$
                \State $\bm{\Sigma}_c = \mathrm{Covariance}(\bm{X}_c)$
                \State $(\bm{\Sigma}_c)_s = \mathrm{Shrinkage}\big(\mathrm{Shrinkage}(\bm{\Sigma}_c)\big)$ 
                \State $(\tilde{\bm{\Sigma}}_c)_s = \mathrm{Normalize} (\bm{\Sigma}_c)_s$
            \EndFor
            \For{$\bm{x} \in \bm{X}_{test}^t$}
                \For{$i \in \lbrace 1, 2, ..., t \rbrace$}
                    \State $\bm{\phi}_{i}(\bm{x}) = f\big(\bm{x}, \mathcal{H}(\bm{e}_i, \Phi) \big)$ 
                    \For{$c \in \bm{\hat{Y}}_{train}^t$}
                        \State $d_M \left( \bm{\phi}_i(\bm{x}), \bm{\mu}_c \right) = \big( \frac{\bm{\phi}_i(\bm{x})}{\lVert \bm{\phi}_i(\bm{x}) \rVert_2} - \frac{\bm{\mu}_c}{\lVert \bm{\mu}_c \rVert_2} \big)^\top (\tilde{\bm{\Sigma}}_c)^{-1}_s \big( \frac{\bm{\phi}_i(\bm{x})}{\lVert \bm{\phi}_i(\bm{x}) \rVert_2} - \frac{\bm{\mu}_c}{\lVert \bm{\mu}_c \rVert_2} \big)$
                    \EndFor
                \EndFor
                \State $c^{\star} = \argmin\limits_{c=1,2,...,|\bm{\mu}|} \bigg( \min\limits_{i=1,2,...,t} d_M \left( \bm{\phi}_i(\bm{x}), \bm{\mu}_c \right) \bigg)$
            \EndFor
        \EndFor    	
    \end{algorithmic}
    \end{algorithm}

\section{Appendix: Stability of \our{} model}\label{app:ablation}

\our{} model has a similar number of hyperparameters as HNET. The most critical ones are $\beta$ and $\lambda$, which control regularization strength. This method also has a hyperparameter $p$ describing the level of zeros in consecutive layers of the semi-binary mask. Moreover, we have to decide whether we want to train the target network or not. For the fixed target network setting $\lambda$ hyperparameter is redundant. We also have to decide whether or not $L_1$ loss is additionally multiplied by the hypernetwork-generated mask values. Furthermore, similarly to HNET, there exists another branch of hyperparameters regarding the networks' shape, for instance, the hypernetwork embedding size, the number of hidden layers and the number of neurons in consecutive layers. Similarly, we have to define the setting of the target network.

\subsection{Loss components}
Hypernetwork regularization using previously trained embeddings is a classical approach in training hypernetwork models, while $L_1$ regularization of the target model is an often used approach in CL. The first component is responsible for the regularization of the hypernetwork weights producing masks, while the second one accounts for the trainable target network weights in \our{}-T. Therefore, both parts are necessary for \our{}-T because it consists of two trainable networks, and we have to prevent radical changes in both sets of weights. In HNET~\cite{von2019continual}, hypernetworks directly produce the target model’s weights, and additional regularization is unnecessary because only one model is trained. When we do not regularize the hypernetwork or main model, the weights are changed too drastically and catastrophic forgetting occurs. In Table ~\ref{tab:loss_components}, we present an ablation study for different hyperparameter settings responsible for the regularization strength (3 runs per one setup) on the Permuted MNIST dataset with 10 tasks using the same architecture, for which we presented results in Table~\ref{tab:continual} in our paper. The results confirm the necessity of both components.

\ctable[
caption = {Mean accuracy with standard deviations for different \our{}+T configurations in the Permuted MNIST dataset.},
label = tab:loss_components,
mincapwidth = \textwidth,
doinside = {\scriptsize},
star = False,
pos = h
]{llc}{}{\FL
	$\mathbf{\beta}$ & $\mathbf{\lambda}$ & mean accuracy (with std. dev.) \ML
    $0.01$ & $0$ & $\mathbf{84.18 \pm 0.18}$ \\
    $0.1$ & $0$ & $80.31 \pm 1.11$ \\
    $1$ & $0$ & $73.68 \pm 0.31$ \\
    $0$ & $0.01$ & $37.82 \pm 1.72$ \\
    $0$ & $0.1$ & $46.32 \pm 1.07$ \\
    $0$ & $1$ & $11.35 \pm 0.00$ \LL
}

\subsection{Permuted MNIST}
We considered different sparsity parameter $p$ values for the optimal \our{}+T found for the Permuted MNIST dataset. The results of these experiments are depicted in Table~\ref{tab:permuted_sparsity} which contains mean scores for 3 training runs per each setup. One can conclude that too intense network sparsification deteriorates the results, although the differences in accuracy are not enormous. It is also necessary to emphasize that, in general, the choice of $p$ is strongly related to the target architecture and the considered dataset. 

\ctable[
caption = {Mean accuracy with standard deviations for sparsity parameter $p$ values for 3 training runs of \our{}+T on the Permuted MNIST dataset. The remaining hyperparameters were the same as in the main experiments, i.e. the embedding size of 24, a hypernetwork with two hidden layers having 100 neurons, a learning rate of $0.001$, $\beta$ of $0.0005$, and masked $L_1$ regularization.},
label = tab:permuted_sparsity,
mincapwidth = \textwidth,
doinside = {\scriptsize},
star = False,
pos = h
]{lc}{}{\FL
	$p$ & mean accuracy (with std. dev.) \ML
    $0$ & $97.551 \pm 0.12$ \\
    $\mathbf{15}$ & $\mathbf{97.680 \pm 0.06}$ \\
    $30$ & $97.596 \pm 0.03$ \\
    $50$ & $97.493 \pm 0.11$ \\
    $70$ & $97.140 \pm 0.14$ \LL
}

\subsection{Split MNIST}
\begin{figure}[!ht]
    \centering
        \begin{subfigure}{0.4\linewidth}
            \includegraphics[trim={0,2cm, 0,2cm, 0,2cm, 0,2cm},clip,width=\linewidth]{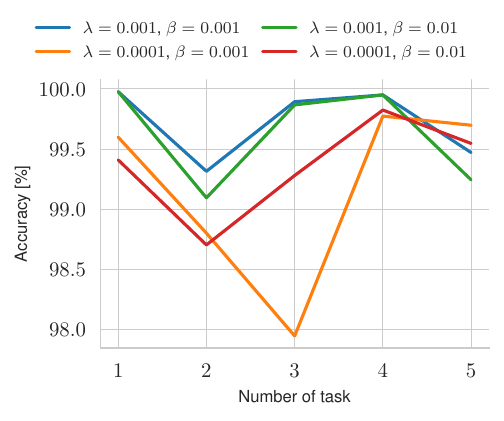}
        \end{subfigure}%
        \begin{subfigure}{0.4\linewidth}
            \includegraphics[trim={0,2cm, 0,2cm, 0,2cm, 0,2cm},clip,width=\linewidth]{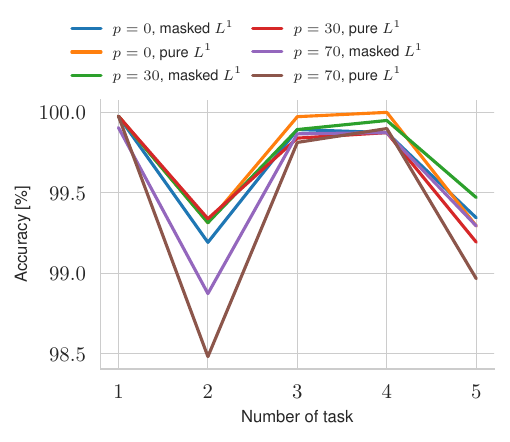}
        \end{subfigure}
        \begin{subfigure}{0.4\linewidth}
            \includegraphics[trim={0,2cm, 0,2cm, 0,2cm, 0,2cm},clip,width=\linewidth]{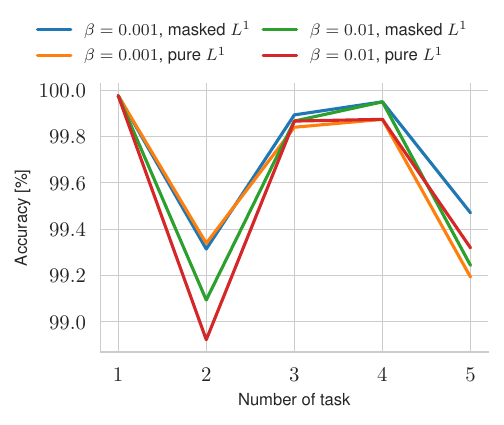}
        \end{subfigure}%
        \begin{subfigure}{0.4\linewidth}
            \includegraphics[trim={0,2cm, 0,2cm, 0,2cm, 0,2cm},clip,width=\linewidth]{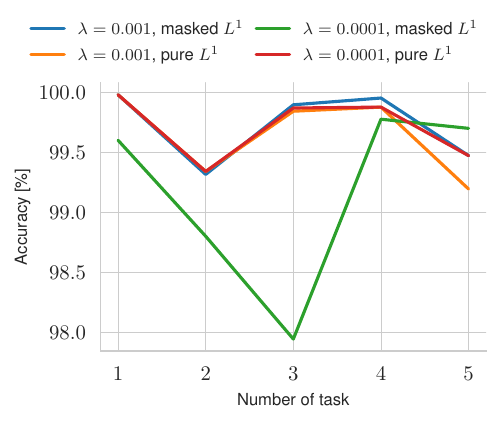}
        \end{subfigure}
        \begin{subfigure}{0.4\linewidth}
            \includegraphics[trim={0,2cm, 0,2cm, 0,2cm, 0,2cm},clip,width=\linewidth]{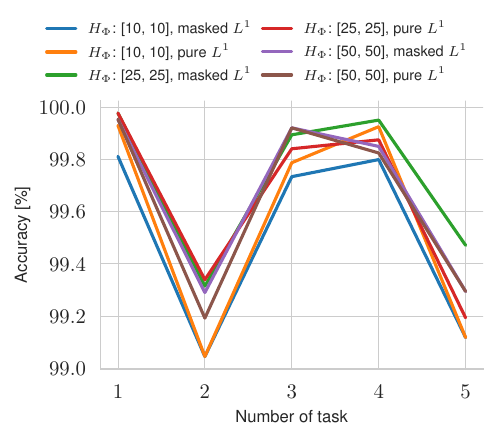}
        \end{subfigure}%
        \begin{subfigure}{0.4\linewidth}
            \includegraphics[trim={0,2cm, 0,2cm, 0,2cm, 0,2cm},clip,width=\linewidth]{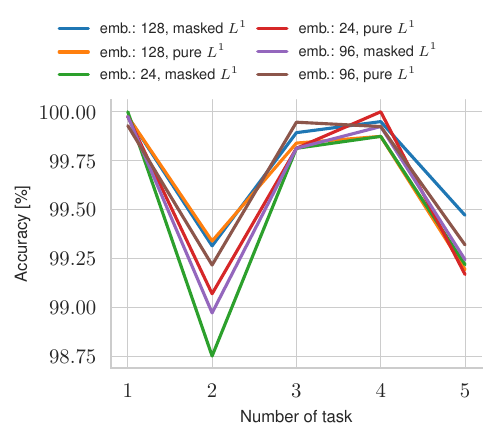}
        \end{subfigure}
    \caption{Mean test accuracy for consecutive CL tasks averaged over two runs of different architecture settings of \our{}+T for five tasks of the Split MNIST dataset. Most of the \our{}+T models achieve the highest classification accuracy for the first CL task while the weakest one for the subsequent task. In many cases, differences in performance of the compared models are slight.  \label{split:stability_analysis}}
\end{figure}

In Fig.~\ref{split:stability_analysis}, we present mean test accuracy for consecutive CL tasks averaged over two runs of different architecture settings of \our{}+T for five tasks of the Split MNIST dataset. Most of the \our{}+T models achieve the highest classification accuracy for the first CL task while the weakest one for the subsequent tasks. In all of the above plots, results are compared with the best hyperparameter setting, i.e. with an embedding size of 128, a hypernetwork having two hidden layers with 25 neurons per each, $\beta = 0.001$, $\lambda = 0.001$, $p = 30$, a batch size of 128 and masked $L^1$ norm. In consecutive subplots, some of the above hyperparameters are changed, and the performance of corresponding models is compared with the most efficient setup.

\subsection{Split CIFAR-100}
The accuracy results of consecutive CL tasks for different settings of \our{} for the Split CIFAR-100 are depicted in the main work, in Fig.~\ref{cifar:ablation}. The left-hand side plot presents single-run training scores for various embedding sizes and $\beta$ regularization parameters for a fixed target network variant. Also, the right-hand side plot shows similar results for two training runs (in terms of means and 95\% confidence intervals) with the trainable target network and different $\beta$ values. One can observe that, in this case, \our{}-T suffers from catastrophic forgetting more than \our{}-F, despite high values of $\beta$. In \our{}-T, $\beta = 0.1$ leads to the highest scores on the first six tasks. Also, when the target model is fixed during training, too intense regularization deteriorates the results: the performance with $\beta = 10$ is the worst among the presented ones.

\subsection{Tiny ImageNet}
    Tables~\ref{tab:ablation_Tiny_fixed_target}--\ref{tab:ablation_Tiny_trainable_target} describe results for different settings of \our{} averaged over forty tasks of the Tiny ImageNet dataset. The results concern mean test accuracy for a single training run. They are divided according to the target network training mode (fixed or trainable) and values of sparsity parameter $p$. All remaining hyperparameters are the same, including a learning rate of $10^{-4}$ and a batch size of $16$.

    When the target network was fixed, strong sparsifying, i.e., removing 50 or 70\% of the weights, was the highest-performing option. Also, a strong regularization, i.e. $\beta$ equal to 1 or even 10, led to better results. For these settings, a larger embedding size and a hypernetwork having one or two hidden layers with 200 neurons were necessary. For smaller values of $p$, a lower embedding size yielded competitive results. Too intense sparsification or too small hypernetwork led to weak scores. These observations are mostly correct for the trainable target network mode. However, in this case, a regularization of the target network was inevitable. Worse solutions were related to the low embedding size and the low strength of the hypernetwork output and target network regularization. Also, no sparsification or too intense pruning was not favourable.

    Except that, we performed experiments with higher learning rate values but $10^{-4}$ was the most promising option. It is necessary to emphasize that \our{} delivers stable results, and for Tiny ImageNet more settings guarantee competitive scores.

\begin{table*}[t]
    \centering
    \caption{Ablation study for Tiny ImageNet trained on ResNet-18 and different values of \our{}+F hyperparameters, i.e. embedding sizes, hypernetwork settings, $\beta$ regularization strength and sparsity parameter $p$. The results are presented as mean task accuracy averaged over 40 tasks after finishing the single training run with a learning rate of $10^{-4}$ and a batch size of $16$. Bold font corresponds to the highest accuracy setting selected for further experiments.}
    \scriptsize
      \begin{tabular}{@{}p{1,9cm}p{1,6cm}>{\centering\arraybackslash}p{1,8cm}>{\raggedleft\arraybackslash}p{0,5cm}>{\centering\arraybackslash}p{0.6cm}p{2cm}}
    	\FL
    \textbf{embedding size} & \textbf{hypernetwork hidden layers} & \textbf{trainable target} & \multicolumn{1}{r}{$\mathbf{\beta}$} & $\mathbf{p}$ & \textbf{mean task accuracy (with std. dev.)} \ML
    $48 $ & $ [200] $ & \xmark & $ 10 $ & $ 0 $ & $ 75.51 \pm 8.12$ \\
    $128 $ & $ [200] $ & \xmark & $ 1 $ & $ 0 $ & $ 75.11 \pm 7.87$ \\
    $48 $ & $ [200] $ & \xmark & $ 0.1 $ & $ 0 $ & $ 75.09 \pm 8.64$ \\
    $256 $ & $ [200] $ & \xmark & $ 100 $ & $ 0 $ & $ 75.09 \pm 7.72$ \\
    $128 $ & $ [200] $ & \xmark & $ 0.1 $ & $ 0 $ & $ 74.95 \pm 7.84$ \\
    $256 $ & $ [200] $ & \xmark & $ 10 $ & $ 0 $ & $ 74.95 \pm 8.18$ \\
    $256 $ & $ [200] $ & \xmark & $ 1 $ & $ 0 $ & $ 74.75 \pm 8.45$ \\
    $48 $ & $ [100] $ & \xmark & $ 10 $ & $ 0 $ & $ 74.74 \pm 8.58$ \\
    $48 $ & $ [100] $ & \xmark & $ 1 $ & $ 0 $ & $ 74.71 \pm 8.28$ \\
    $48 $ & $ [200,200] $ & \xmark & $ 1 $ & $ 0 $ & $ 74.28 \pm 8.01$ \\
    $256 $ & $ [200] $ & \xmark & $ 0.1 $ & $ 0 $ & $ 74.24 \pm 8.88$ \\
    $48 $ & $ [200] $ & \xmark & $ 1 $ & $ 0 $ & $ 74.19 \pm 8.45$ \\
    $48 $ & $ [100] $ & \xmark & $ 0.1 $ & $ 0 $ & $ 73.98 \pm 8.30$ \\
    $48 $ & $ [200,200] $ & \xmark & $ 0.1 $ & $ 0 $ & $ 73.64 \pm 7.86$ \ML

    $256 $ & $ [200] $ & \xmark & $ 1$ & $ 30 $ & $ 75.82 \pm 8.03$ \\
    $256 $ & $ [200] $ & \xmark & $ 10 $ & $ 30 $ & $ 75.44 \pm 7.87$ \\
    $256 $ & $ [200,200] $ & \xmark & $ 10 $ & $ 30 $ & $ 75.37 \pm 7.56$ \\
    $128 $ & $ [200] $ & \xmark & $ 10 $ & $ 30 $ & $ 75.35 \pm 7.83$ \\
    $128 $ & $ [200] $ & \xmark & $ 0.1 $ & $ 30 $ & $ 75.34 \pm 8.33$ \\
    $256 $ & $ [200] $ & \xmark & $ 0.1 $ & $ 30 $ & $ 75.17 \pm 7.61$ \\ 
    $256 $ & $ [200] $ & \xmark & $ 100 $ & $ 30 $ & $ 75.13 \pm 7.97$ \\
    $48 $ & $ [200] $ & \xmark & $ 1 $ & $ 30 $ & $ 75.02 \pm 8.19$ \\
    $48 $ & $ [200,200] $ & \xmark & $ 1 $ & $ 30 $ & $ 74.78 \pm 8.18$ \\
    $48 $ & $ [200] $ & \xmark & $ 0.1 $ & $ 30 $ & $ 74.67 \pm 7.47$ \\
    $128 $ & $ [200] $ & \xmark & $ 1 $ & $ 30 $ & $ 74.64 \pm 8.57$ \\
    $256 $ & $ [200,200] $ & \xmark & $ 1 $ & $ 30 $ & $ 74.55 \pm 8.26$ \\
    $48 $ & $ [100] $ & \xmark & $ 10 $ & $ 30 $ & $ 74.26 \pm 8.48$ \\
    $256 $ & $ [200,200] $ & \xmark & $ 100 $ & $ 30 $ & $ 74.22 \pm 9.04$ \\
    $48 $ & $ [100] $ & \xmark & $ 1 $ & $ 30 $ & $ 74.13 \pm 8.71$ \\
    $256 $ & $ [200,200] $ & \xmark & $ 0.1 $ & $ 30 $ & $ 74.01 \pm 7.61$ \\ 
    $48 $ & $ [200,200] $ & \xmark & $ 0.1 $ & $ 30 $ & $ 73.26 \pm 8.33$ \\
    $48 $ & $ [100] $ & \xmark & $ 0.1 $ & $ 30 $ & $ 73.13 \pm 8.18$ \ML

    $256 $ & $ [200] $ & \xmark & $ 10 $ & $ 50 $ & $ 75.94 \pm 8.35$ \\
    $128 $ & $ [200] $ & \xmark & $ 10 $ & $ 50 $ & $ 75.93 \pm 7.40$ \\
    $256 $ & $ [200] $ & \xmark & $ 1 $ & $ 50 $ & $ 75.92 \pm 7.90$ \\
    $256 $ & $ [200,200] $ & \xmark & $ 10 $ & $ 50 $ & $ 75.71 \pm 8.17$ \\
    $128 $ & $ [200] $ & \xmark & $ 1 $ & $ 50 $ & $ 75.67 \pm 8.01$ \\
    $256 $ & $ [200,200] $ & \xmark & $ 1 $ & $ 50 $ & $ 75.40 \pm 8.01$ \\
    $128 $ & $ [200] $ & \xmark & $ 0.1 $ & $ 50 $ & $ 75.10 \pm 8.11$ \\
    $256 $ & $ [200] $ & \xmark & $ 0.1 $ & $ 50 $ & $ 74.99 \pm 7.98$ \\
    $48 $ & $ [200,200] $ & \xmark & $ 1 $ & $ 50 $ & $ 74.72 \pm 7.84$ \\
    $256 $ & $ [200,200] $ & \xmark & $ 0.1 $ & $ 50 $ & $ 74.58 \pm 7.65$ \\
    $48 $ & $ [100] $ & \xmark & $ 1 $ & $ 50 $ & $ 74.41 \pm 7.97$ \\
    $48 $ & $ [100] $ & \xmark & $ 10 $ & $ 50 $ & $ 74.30 \pm 8.39$ \\
    $48 $ & $ [200] $ & \xmark & $ 0.1 $ & $ 50 $ & $ 74.02 \pm 8.97$ \\
    $256 $ & $ [200] $ & \xmark & $ 100 $ & $ 50 $ & $ 73.96 \pm 9.31$ \\
    $48 $ & $ [200,200] $ & \xmark & $ 0.1 $ & $ 50 $ & $ 73.52 \pm 8.33$ \\
    $256 $ & $ [200,200] $ & \xmark & $ 100 $ & $ 50 $ & $ 73.40 \pm 9.74$ \\
    $48 $ & $ [100] $ & \xmark & $ 0.1 $ & $ 50 $ & $ 72.60 \pm 8.09$ \ML

    $\mathbf{256} $ & $\mathbf{[200,200]} $ & \xmark & $ \mathbf{1} $ & $\mathbf{70}$ & $ \mathbf{76.04 \pm 7.83}$ \\
    $256 $ & $ [200,200] $ & \xmark & $ 10 $ & $ 70 $ & $ 75.78 \pm 7.58$ \\
    $256 $ & $ [200] $ & \xmark & $ 1 $ & $ 70 $ & $ 75.59 \pm 8.09$ \\
    $48 $ & $ [200,200] $ & \xmark & $ 1 $ & $ 70 $ & $ 75.09 \pm 8.06$ \\
    $256 $ & $ [200] $ & \xmark & $ 0.1 $ & $ 70 $ & $ 74.86 \pm 7.25$ \\
    $48 $ & $ [200] $ & \xmark & $ 0.1 $ & $ 70 $ & $ 74.68 \pm 7.77$ \\
    $256 $ & $ [200] $ & \xmark & $ 10 $ & $ 70 $ & $ 74.66 \pm  8.17$ \\
    $256 $ & $ [200,200] $ & \xmark & $ 0.1 $ & $ 70 $ & $ 74.59 \pm 7.89$ \\
    $48 $ & $ [100] $ & \xmark & $ 1 $ & $ 70 $ & $ 73.52 \pm 8.31$ \\
    $48 $ & $ [200,200] $ & \xmark & $ 0.1 $ & $ 70 $ & $ 73.29 \pm 7.79$ \\
    $256 $ & $ [200,200] $ & \xmark & $ 100 $ & $ 70 $ & $ 72.33 \pm 10.41$ \\
    $48 $ & $ [100] $ & \xmark & $ 0.1 $ & $ 70 $ & $ 71.96 \pm 8.08$ \ML

    $256 $ & $ [200,200] $ & \xmark & $ 1 $ & $ 90 $ & $ 73.06 \pm 8.31$ \\
    $256 $ & $ [200] $ & \xmark & $ 0.1 $ & $ 90 $ & $ 72.92 \pm 8.88$ \\
    $256 $ & $ [200] $ & \xmark & $ 1 $ & $ 90 $ & $ 72.76 \pm 10.22$ \\
    $256 $ & $ [200] $ & \xmark & $ 10 $ & $ 90 $ & $ 72.70 \pm 8.88$ \\
    $256 $ & $ [200,200] $ & \xmark & $ 0.1 $ & $ 90 $ & $ 72.56 \pm  8.36$ \\
    $256 $ & $ [200,200] $ & \xmark & $ 10 $ & $ 90 $ & $ 72.04 \pm 8.97$ \\
    $256 $ & $ [200,200] $ & \xmark & $ 100 $ & $ 90 $ & $ 67.60 \pm 12.17$ \LL
    
     \end{tabular}
    \label{tab:ablation_Tiny_fixed_target}
\end{table*}

\begin{table*}[t]
    \centering
    \caption{Ablation study for Tiny ImageNet trained on ResNet-18 and different values of \our{}+T hyperparameters, i.e. embedding sizes, hypernetwork settings, $\beta$ and $\lambda$ regularizations strength and sparsity parameter $p$. The results are presented as mean task accuracy averaged over 40 tasks after finishing the single training run with a learning rate of $10^{-4}$ and a batch size of $16$. Bold font corresponds to the model with the highest accuracy while italic font denotes the model setting chosen for main experiments with 5 training runs, due to higher accuracy with multiple runs.}
    \scriptsize
      \begin{tabular}{@{}p{1,9cm}p{1,6cm}>{\centering\arraybackslash}p{1,8cm}>{\raggedleft\arraybackslash}p{0,5cm}>{\centering\arraybackslash}p{0.6cm}>{\raggedleft\arraybackslash}p{0,5cm}p{2cm}}
    	\FL
    \textbf{embedding size} & \textbf{hypernetwork hidden layers} & \textbf{trainable target} & \multicolumn{1}{r}{$\mathbf{\beta}$} & $\mathbf{p}$ & \multicolumn{1}{r}{$\mathbf{\lambda}$} & \textbf{mean task accuracy (with std. dev.)} \ML
    $48 $ & $ [200] $ & \checkmark & $ 0.1 $ & $ 0 $ & $ 1 $ & $ 75.00 \pm 7.77$ \\
    $48 $ & $ [200] $ & \checkmark & $ 1 $ & $ 0 $ & $ 0.1 $ & $ 74.68 \pm 7.26$ \\
    $48 $ & $ [200] $ & \checkmark & $ 1 $ & $ 0 $ & $ 1 $ & $ 74.56 \pm 8.26$ \\
    $48 $ & $ [200] $ & \checkmark & $ 10 $ & $ 0 $ & $ 1 $ & $ 74.29 \pm 7.86$ \\
    $48 $ & $ [200,200] $ & \checkmark & $ 10 $ & $ 0 $ & $ 0.1 $ & $ 74.28 \pm 8.41$ \\
    $48 $ & $ [200] $ & \checkmark & $ 10 $ & $ 0 $ & $ 0.01 $ & $ 74.24 \pm 8.91$ \\
    $48 $ & $ [200] $ & \checkmark & $ 10 $ & $ 0 $ & $ 0.1 $ & $ 74.24 \pm 8.36$ \\
    $48 $ & $ [200,200] $ & \checkmark & $ 10 $ & $ 0 $ & $ 1 $ & $ 74.04 \pm 8.20$ \\
    $48 $ & $ [200] $ & \checkmark & $ 0.1 $ & $ 0 $ & $ 0.1 $ & $ 73.89 \pm 8.49$ \\
    $48 $ & $ [200] $ & \checkmark & $ 0.1 $ & $ 0 $ & $ 0.01 $ & $ 73.50 \pm 8.28$ \\
    $48 $ & $ [200,200] $ & \checkmark & $ 1 $ & $ 0 $ & $ 1 $ & $ 73.48 \pm 7.88$ \\
    $48 $ & $ [200,200] $ & \checkmark & $ 10 $ & $ 0 $ & $ 0.01 $ & $ 73.48 \pm 8.56$ \\
    $48 $ & $ [200,200] $ & \checkmark & $ 1 $ & $ 0 $ & $ 0.01 $ & $ 73.39 \pm 8.4$ \\
    $48 $ & $ [200,200] $ & \checkmark & $ 0.1 $ & $ 0 $ & $ 1 $ & $ 73.27 \pm 7.6$ \\
    $48 $ & $ [200,200] $ & \checkmark & $ 1 $ & $ 0 $ & $ 0.1 $ & $ 73.26 \pm 7.77$ \\
    $48 $ & $ [200,200] $ & \checkmark & $ 0.1 $ & $ 0 $ & $ 0.01 $ & $ 73.17 \pm 8.32$ \\
    $48 $ & $ [200] $ & \checkmark & $ 1 $ & $ 0 $ & $ 0.01 $ & $ 73.10 \pm 8.22$ \\
    $48 $ & $ [200,200] $ & \checkmark & $ 0.1 $ & $ 0 $ & $ 0.1 $ & $ 72.41 \pm 8.91$ \ML

    $\mathit{256} $ & $ \mathit{[200,200]} $ & \checkmark & $ \mathit{10} $ & $ \mathit{15} $ & $ \mathit{0.1} $ & $ \mathit{75.14 \pm 7.67}$ \\
    $256 $ & $ [200,200] $ & \checkmark & $ 10 $ & $ 15 $ & $ 0.01 $ & $ 74.39 \pm  8.37$ \\
    $48 $ & $ [200] $ & \checkmark & $ 0.1 $ & $ 15 $ & $ 10 $ & $ 74.37 \pm 7.70$ \\
    $256 $ & $ [200,200] $ & \checkmark & $ 1 $ & $ 15 $ & $ 0.1 $ & $ 74.31 \pm  8.73$ \\
    $256 $ & $ [200,200] $ & \checkmark & $ 10 $ & $ 15 $ & $ 1 $ & $ 74.29 \pm  7.97$ \\
    $256 $ & $ [200,200] $ & \checkmark & $ 1 $ & $ 15 $ & $ 1 $ & $ 74.08 \pm  8.26$ \\
    $256 $ & $ [200,200] $ & \checkmark & $ 1 $ & $ 15 $ & $ 0.01 $ & $ 73.80 \pm  7.78$ \\
    $48 $ & $ [200] $ & \checkmark & $ 0.01 $ & $ 15 $ & $ 10 $ & $ 73.52 \pm  8.30$ \\
    $48 $ & $ [200] $ & \checkmark & $ 0.1 $ & $ 15 $ & $ 1 $ & $ 73.45 \pm 7.86$ \\
    $48 $ & $ [200] $ & \checkmark & $ 0.01 $ & $ 15 $ & $ 1 $ & $ 73.20 \pm 8.46$ \ML

    $256 $ & $ [200,200] $ & \checkmark & $ 10 $ & $ 30 $ & $ 1 $ & $ 75.04 \pm 8.72$ \\
    $256 $ & $ [200,200] $ & \checkmark & $ 10 $ & $ 30 $ & $ 0.1 $ & $ 75.02 \pm 7.66$ \\
    $256 $ & $ [200,200] $ & \checkmark & $ 10 $ & $ 30 $ & $ 0.01 $ & $ 74.90 \pm 8.08$ \\
    $256 $ & $ [200,200] $ & \checkmark & $ 1 $ & $ 30 $ & $ 1 $ & $ 74.89 \pm 7.78$ \\
    $256 $ & $ [200,200] $ & \checkmark & $ 1 $ & $ 30 $ & $ 0.1 $ & $ 74.71 \pm 8.07$ \\
    $256 $ & $ [200,200] $ & \checkmark & $ 10 $ & $ 30 $ & $ 0.1 $ & $ 74.33 \pm 8.06$ \\
    $256 $ & $ [200,200] $ & \checkmark & $ 10 $ & $ 30 $ & $ 0.01 $ & $ 74.00 \pm 8.28$ \\
    $256 $ & $ [200,200] $ & \checkmark & $ 1 $ & $ 30 $ & $ 0.01 $ & $ 73.91 \pm 8.57$ \ML

    $\mathbf{256}$ & $\mathbf{[200,200]} $ & \checkmark & $\mathbf{1} $ & $\mathbf{50} $ & $\mathbf{1} $ & $\mathbf{75.21 \pm 8.19}$ \\
    $256 $ & $ [200,200] $ & \checkmark & $ 10 $ & $ 50 $ & $ 0.1 $ & $ 74.57 \pm 8.16$ \\
    $256 $ & $ [200,200] $ & \checkmark & $ 1 $ & $ 50 $ & $ 0.1 $ & $ 74.42 \pm 8.01$ \\
    $256 $ & $ [200,200] $ & \checkmark & $ 1 $ & $ 50 $ & $ 0.01 $ & $ 74.36 \pm 8.16$ \\
    $256 $ & $ [200,200] $ & \checkmark & $ 10 $ & $ 50 $ & $ 0.01 $ & $ 74.26 \pm 8.32$ \\
    $256 $ & $ [200,200] $ & \checkmark & $ 1 $ & $ 50 $ & $ 0.01 $ & $ 74.14 \pm 8.09$ \\
    $256 $ & $ [200,200] $ & \checkmark & $ 1 $ & $ 50 $ & $ 0.1 $ & $ 74.14 \pm 8.17$ \\
    $48 $ & $ [200] $ & \checkmark & $ 0.1 $ & $ 50 $ & $ 10 $ & $ 73.32 \pm 8.92$ \\
    $48 $ & $ [200] $ & \checkmark & $ 0.1 $ & $ 50 $ & $ 1 $ & $ 73.27 \pm 8.09$ \\
    $48 $ & $ [200] $ & \checkmark & $ 0.01 $ & $ 50 $ & $ 10 $ & $ 72.28 \pm 8.62$ \\
    $48 $ & $ [200] $ & \checkmark & $ 0.01 $ & $ 50 $ & $ 1 $ & $ 72.26 \pm 8.88$ \ML
    
    $256 $ & $ [200,200] $ & \checkmark & $ 1 $ & $ 70 $ & $ 0.1 $ & $ 75.13 \pm 7.49$ \\
    $256 $ & $ [200,200] $ & \checkmark & $ 1 $ & $ 70 $ & $ 1 $ & $ 74.74 \pm 8.50$ \\
    $256 $ & $ [200,200] $ & \checkmark & $ 1 $ & $ 70 $ & $ 0.1 $ & $ 74.60 \pm 8.44$ \\
    $256 $ & $ [200,200] $ & \checkmark & $ 10 $ & $ 70 $ & $ 0.1 $ & $ 74.45 \pm  8.43$ \\
    $256 $ & $ [200,200] $ & \checkmark & $ 1 $ & $ 70 $ & $ 0.01 $ & $ 74.01 \pm 8.11$ \\
    $256 $ & $ [200,200] $ & \checkmark & $ 1 $ & $ 70 $ & $ 0.01 $ & $ 73.96 \pm  8.04$ \\
    $256 $ & $ [200,200] $ & \checkmark & $ 10 $ & $ 70 $ & $ 0.01 $ & $ 73.77 \pm  7.85$ \\
    $48 $ & $ [200] $ & \checkmark & $ 0.1 $ & $ 70 $ & $ 1 $ & $ 73.57 \pm 8.24$ \\
    $48 $ & $ [200] $ & \checkmark & $ 0.1 $ & $ 70 $ & $ 10 $ & $ 73.30 \pm 8.27$ \\
    $48 $ & $ [200] $ & \checkmark & $ 0.01 $ & $ 70 $ & $ 1 $ & $ 71.71 \pm 8.23$ \\
    $48 $ & $ [200] $ & \checkmark & $ 0.01 $ & $ 70 $ & $ 10 $ & $ 71.10 \pm 9.20$ \LL
    \end{tabular}
    \label{tab:ablation_Tiny_trainable_target}
\end{table*}

\section{Appendix: Time consumption}
We depicted in Table~\ref{tab:calculation} the mean training time of \our{}+T for five tasks of the Split MNIST, ten tasks of Permuted MNIST, ten tasks (with ten classes in each of them) or five tasks (with twenty classes in each of them) in the case of Split CIFAR-100 and forty tasks of Tiny ImageNet. The calculations were performed using NVIDIA GeForce RTX 2070, 2080 Ti, 3080 and 4090 graphic cards. For the easiest dataset, \our{}+T needs only slightly more than 21 minutes. In the case of Permuted MNIST, which consists of more advanced CL tasks, \our{} needs less than 2 hours. For Split CIFAR-100 and a more complicated convolutional architecture, computation times were less than two hours for the ten-task scenario and about three and a half hours for the five-task scenario. It is necessary to emphasize that each task from CIFAR-100 was trained through 200 epochs.

Another dataset, i.e. Tiny ImageNet, contains images having larger spatial resolution than CIFAR-100 ($64$ instead of $32$). Each of the 40 five-class CL tasks was learned through 10 epochs. The total computation time was equal to around four hours. It would be much shorter if 
the time necessary for loading the dataset and creating 40 tasks would not be considered (several dozen minutes).

\ctable[
caption = {Mean training time of \our{}+T for different datasets.},
label = tab:calculation,
mincapwidth = \textwidth,
doinside = {\footnotesize},
star = False,
pos = h
]{ll}{}{\FL
	\textbf{Dataset} & \textbf{Mean calculation time in HH:MM:SS (with std. dev.)} \ML
    Split MNIST & 00:21:06 $\pm$ 00:02:37 \\
    Permuted MNIST & 01:45:14 $\pm$ 00:04:07 \\
    Split CIFAR-100 (10 tasks, 10 classes each) & 01:47:25 $\pm$ 00:02:51 \\
    Split CIFAR-100 (5 tasks, 20 classes each) & 03:35:04 $\pm$ 00:03:23 \\
    Tiny ImageNet & 04:00:02 $\pm$ 00:17:38 \LL
}

\end{document}